\pdfoutput=1

\documentclass[11pt]{article}

\usepackage[preprint]{acl}

\usepackage{times}
\usepackage{latexsym}
\usepackage[normalem]{ulem}
\usepackage{makecell}
\usepackage{siunitx}
\usepackage{tabularx}
\usepackage{xcolor,colortbl}
\usepackage{pifont}
\usepackage{booktabs}
\usepackage{multirow}
\usepackage{multicol}
\usepackage{graphicx}
\usepackage{amsmath}
\usepackage{amsthm}
\usepackage{amssymb}
\usepackage{mathtools}
\usepackage{amsfonts}
\usepackage{bbm}
\usepackage{bm}
\usepackage{soul}
\usepackage{enumitem}
\usepackage{stfloats}
\usepackage[T1]{fontenc}
\usepackage[utf8]{inputenc}
\usepackage{microtype}
\usepackage{inconsolata}

\definecolor{mybluebox}{HTML}{006FEB}
\newcommand{\brtext}[1]{{\color[HTML]{BE4F28} {#1}}}
\newcommand{\grtext}[1]{{\color[HTML]{008B4F} {#1}}}
\newcommand{\bltext}[1]{{\color[HTML]{006EF4} {#1}}}
\newcommand{\blueboxword}[1]{\setlength{\fboxrule}{1pt}\setlength{\fboxsep}{1pt}\fcolorbox{mybluebox}{white}{#1}}

\title{Unveiling the Lexical Sensitivity of LLMs: Combinatorial Optimization for Prompt Enhancement}

\author{
  Pengwei Zhan$^{\diamondsuit}$$^{\clubsuit}$,
  Zhen Xu$^{\diamondsuit}$,
  Qian Tan$^{\diamondsuit}$\thanks{\ Corresponding Author.},
  Jie Song$^{\diamondsuit}$$^{\clubsuit}$,
  Ru Xie$^{\diamondsuit}$$^{\clubsuit}$ \\
  $^{\diamondsuit}$Institute of Information Engineering, Chinese Academy of Sciences, Beijing, China \\
  $^{\clubsuit}$School of Cyber Security, University of Chinese Academy of Sciences, Beijing, China \\
  \texttt{\{zhanpengwei,xuzhen,tanqian,songjie,xieru\}@iie.ac.cn}}

\begin{document}
\maketitle
\begin{abstract}

Large language models (LLMs) demonstrate exceptional instruct-following ability to complete various downstream tasks. Although this impressive ability makes LLMs flexible task solvers, their performance in solving tasks also heavily relies on instructions. In this paper, we reveal that LLMs are over-sensitive to lexical variations in task instructions, even when the variations are imperceptible to humans. By providing models with neighborhood instructions, which are closely situated in the latent representation space and differ by only one semantically similar word, the performance on downstream tasks can be vastly different. Following this property, we propose a black-box \textbf{C}ombinatorial \textbf{O}ptimization framework for \textbf{P}rompt \textbf{L}exical \textbf{E}nhancement (COPLE). COPLE performs iterative lexical optimization according to the feedback from a batch of proxy tasks, using a search strategy related to word influence. Experiments show that even widely-used human-crafted prompts for current benchmarks suffer from the lexical sensitivity of models, and COPLE recovers the declined model ability in both instruct-following and solving downstream tasks.

\end{abstract}

\section{Introduction}

Language models have achieved remarkable performance in recent years, particularly those referred to as large language models (LLMs), which contain scaled-up parameters and size~\cite{Scaling2020Kaplan,Brown2020GPT3,Training2022Hoffmann,Openai2022ChatGPT,Touvron2023Llama2,Jiang2023Mistral}. These models demonstrate an exceptional ability to follow human instructions and complete downstream tasks after instruction tuning~\cite{Ouyang2022instructgpt}. In contrast to masked language models (MLMs) like BERT~\cite{devlin2018bert}, LLMs do not require the addition and training of extra layers on top of the pre-trained base model to adapt to different downstream tasks. Instead, they complete a wide range of tasks in the same way of generating text, by following different task instructions.

\begin{figure}[!t]
    \centering
  \includegraphics[width=1.0\columnwidth]{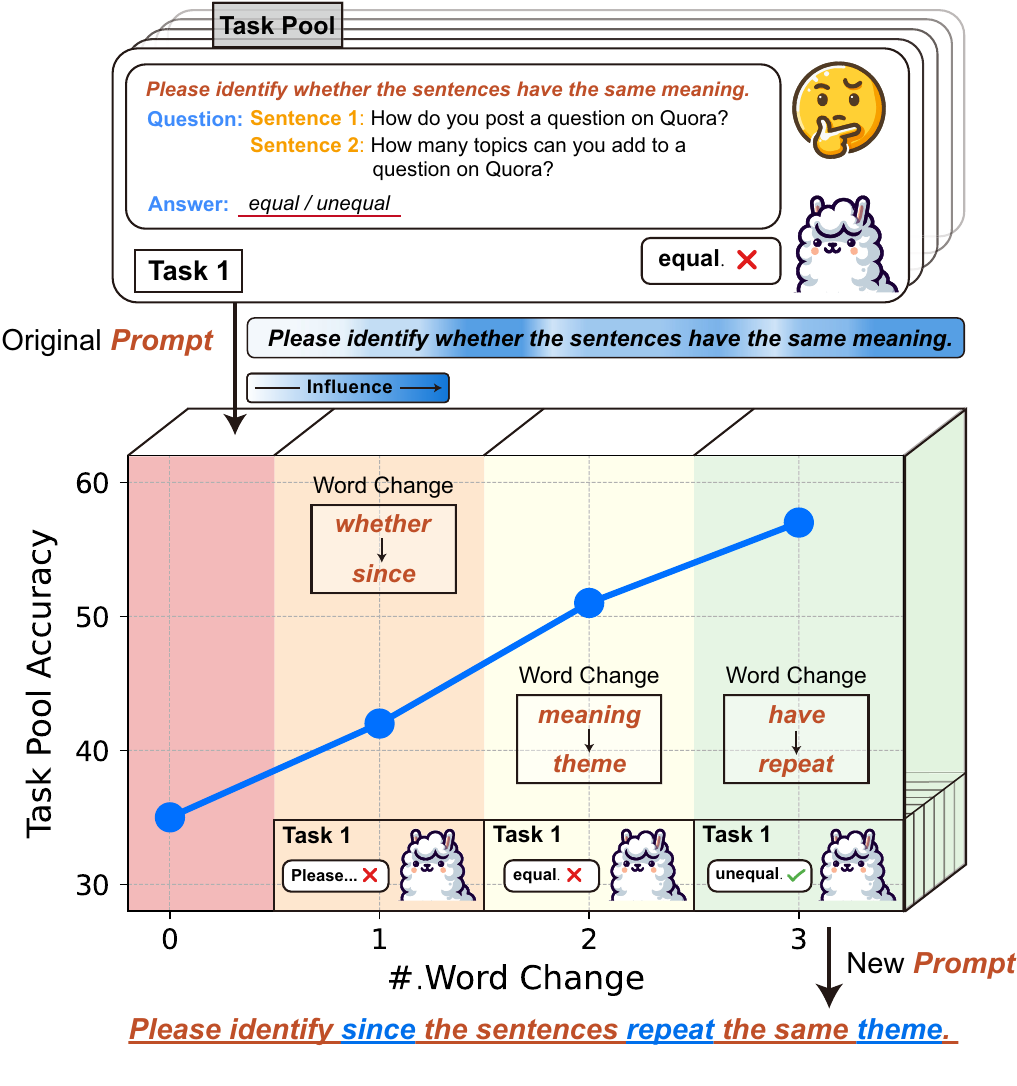}
  \caption{Prompt lexical enhancement from a combinatorial optimization perspective. Initially, we provide the prompt "\emph{Please identify whether the sentences have the same meaning}" for \texttt{Llama-2-7B-chat} to complete the tasks from Quora Question Pairs2 (QQP), and combine the validation set of QQP with the prompt as a predefined task pool, with each example being an individual task. By iteratively substituting the most influential words in the prompt with semantically similar words picked from the potential search space, we find the optimal prompt "\emph{Please identify since the sentences repeat the same theme}" that increases the accuracy from 35\% to 57\%. The details of operations can be found in \S\ref{cople}.}
  \label{overview}
  \vspace{-2mm}
\end{figure}

Although the instruction-following ability of LLMs makes them flexible task solvers, their performance on solving tasks also significantly depends on the instructions (i.e., prompts), which are mainly designed by human intuitively and empirically~\cite{Wei2022CoT,LuBM0S22,KojimaGRMI22,ZhouSHWS0SCBLC23}. These manually designed prompts that incorporated with human knowledge effectively improve the model's performance on specific tasks. However, following \citet{gao2018black}, \citet{garg2020bae} and \citet{feng2018pathologies}, even a minor lexical modification in the input that is imperceptible to humans can lead to vastly different model attention and outputs. Therefore, it is natural to wonder: \ul{\emph{whether the prompts carefully constructed by humans maximize LLMs' performance on downstream tasks?}} For example, in the context of a sentiment classification task, while humans may confidently assert that the prompt "\emph{Please} {\color[HTML]{004DA1}\emph{classify}} \emph{the sentiment of the given text}" outperforms "\emph{Check the given text}", it is hard to say whether it would outperform a prompt like "\emph{Please} {\color[HTML]{004DA1}\emph{analyze}} \emph{the sentiment of the given text}".

The unexpected sensitivity of language models to these imperceptible lexical perturbations suggests the possible existence of an alternate prompt, which is differs from the original prompt by only a few substituted words, yet yields superior performance on downstream tasks. This insight allows us to frame the process of discovering such an optimal prompt as a combinatorial optimization problem~\cite{Blair90}, which consists of two key components: the \emph{search space} and the \emph{search method}. The search space can be defined as the set of all potential substitutions for each word in the original prompt, while the search method specifies the strategy for exploring this space and identifying the optimal substitutions. Figure \ref{overview} provides a more intuitive example of the process of finding the optimal prompt from a lexical combinatorial optimization perspective. We argue that even without the complex prompt engineering, minor lexical modifications to prompts yield substantial improvements to a model's performance.

In this paper, we reveal the notable sensitivity of LLMs to lexical variations in prompts, which potentially undermine the effectiveness of human-crafted prompts, from the view of combinatorial optimization. Based on our findings, we also propose a black-box \textbf{C}ombinatorial \textbf{O}ptimization framework for \textbf{P}rompt \textbf{L}exical \textbf{E}nhancement (COPLE). We summarize our main contributions as follows:

\begin{enumerate}
  \item We intuitively reveal the notable sensitivity of LLMs to the lexical choices in prompts, which suggests the existence of prompts that, while highly similar to the original, can lead to improved performance on downstream tasks.
  \item We propose COPLE, a black-box combinatorial optimization framework that enhances prompts on downstream tasks, which performs iteratively lexical optimization under the guidance of word influence.
  \item We evaluate COPLE on popular datasets, models, and initial prompts. The results show that COPLE effectively maximizes the performance of prompts in downstream tasks with only lexical modifications, without accessing model parameters or involving complex prompt engineering with human participation.
\end{enumerate}

\section{Related Work}

\paragraph{Sensitivity to Imperceptible Changes.}

The outstanding performance of language models seems to be built upon their excellent understanding of text~\cite{devlin2018bert,Dong2019Unified,Radford2018GPT1,Radford2018GPT2,Brown2020GPT3}. However, previous works reveal that even imperceptible input perturbations, which do not affect human comprehension, can lead to significant changes in the model's output~\cite{goodfellow2014explaining,papernot2016limitations,Zhan2022Mitigating,Zhan2023Contrastive,carlini2017towards}. This property has been widely exploited to create adversarial examples, where small modifications to the embedding or input text can cause the model to generate incorrect answers~\cite{gao2018black,Zhan2022PARSE,Zhan2024Rethinking,li2018textbugger,li2020bert,Zhan2023Similarizing,zang2019word}. Therefore, we believe even humans experienced in designing prompts may overlook the performance discrepancies caused by such imperceptible changes.

\paragraph{Prompt Tuning and Optimizing.}

Similarly, recent efforts to optimize prompts for LLMs find that not only the content~\cite{KojimaGRMI22,Yang2023larger} but also the format~\cite{ZhouSHWS0SCBLC23,LuBM0S22,Wei2023larger,Madaan2022text,Prasad2023GrIPS} of the prompt, such as the order of examples and phrases, significantly influence the model performance. Consequently, this sensitivity of language models to minor changes makes the optimal prompt found by the community increasingly complex. For example, \citet{Xu2023ExpertPrompting} transforms a prompt of length 6 into one exceeding 900 tokens. 
However, we argue that also due to this sensitivity, complex variations of the prompt should not be the first operation in prompt optimization, as the performance could be inadvertently constrained by the specific words employed in the prompt.
This proposition distinguishes our work from previous studies on prompt optimization: given a prompt that has proven initially effective, we focus on the lexical influence of the prompt on model performance, attempting to \ul{\emph{recover}} the potential performance drop caused by lexical choices, rather than \ul{\emph{creating}} a prompt that yields optimal results from scratch~\cite{ShinRLWS20,ZhouMHPPCB23,Zhang2022tempera,Yang2023larger,Prasad2023GrIPS}.

\section{Methodology}

\subsection{Prompt Enhancement}

Suppose we are given a data distribution $\mathcal{D} $ over a sequence of downstream tasks $\mathcal{Z} = \{\mathcal{X}, \mathcal{Y} \}$, and each task in the entire task set can be seen as a pair of question and answer $\{\boldsymbol{X}, \boldsymbol{Y} \}$ that both consist of multiple tokens.To recognize a pre-trained autoregressive language model $f_{\boldsymbol{\theta}}$ as the task solver on the task set $\mathcal{Z}$, we hope it can map the input questions to the output answers $f_{\boldsymbol{\theta}}: \mathcal{X} \rightarrow \mathcal{Y}$. However, as model $f_{\boldsymbol{\theta}}$ is not fine-tuned for a specific task, this mapping can only be held with the help of a task-specific prompt $\boldsymbol{P}_{\mathcal{Z}}(\boldsymbol{X}) = (\boldsymbol{D}, \boldsymbol{E}^{\prime}, \boldsymbol{X}, \boldsymbol{V}) $, where $\boldsymbol{D}$ is the task description, $\boldsymbol{E}^{\prime}$ are optional demo examples for few-shot learning, and $\boldsymbol{V}$ is the verbalizer that limits the responses of the model to a set of label words. We can then formulate the performance of the model on the task set as:
\begin{equation}
\label{performance}
        \mathbb{E}_{(\boldsymbol{X}, \boldsymbol{Y}) \sim \mathcal{D}} [\mathcal{L}(f_{\boldsymbol{\theta}}(\boldsymbol{P}_{\mathcal{Z}}(\boldsymbol{X})), \boldsymbol{Y})]
\end{equation}
where $\mathcal{L}$ is a task-specific loss function that measures the discrepancy between the model's output and the ground truth answer. Following this, we can find that directly optimizing the prompt $\boldsymbol{P}_{\mathcal{Z}}(\boldsymbol{X})$ in the discrete token space is challenging due to the non-differentiable nature of text and the large search space.
Therefore, it is more suitable to frame the process of optimizing the prompt as a combinatorial optimization problem, where we aim to find the optimal combination of tokens from a predefined search space that consists of candidate tokens. 

In this paper, to be more specific, we focus on investigating the influence of minor lexical changes on the \emph{task description} part of the prompt. Let $\boldsymbol{D} = (d_1, d_2, \dots, d_n)$ be the sequence of tokens in the task description, and $\mathcal{C}_i \in \mathcal{C}$ denote the search space of token $d_i$. The optimal alternative task description that recovers the potential performance drop caused by wording can thus be formulated as:
\begin{equation}
\begin{aligned}
\label{alternative_task_description}
         \boldsymbol{D}^* &= (d_1^*, d_2^*, \dots, d_n^*), \\
        & \text{s.t. }\quad d_i^* \in \mathcal{C}_i \\
        & \text{and } \ \ \ \forall i\in \{1,\ldots,n\},\ \ \Delta d_i < \delta
\end{aligned}
\end{equation}
where $\Delta d_i$ denotes the difference between $d_i$ and $d_i^*$, and $\delta$ denotes a small maximum allowed difference between them, which limits the possible candidates in the search space to tokens that are semantically similar to the original one. This optimal task description $\boldsymbol{D}^*$ is expected to minimize the expected loss on downstream tasks:
\begin{equation}
\label{combinatorial_optimization}
        \boldsymbol{D}^{*} = \underset{\boldsymbol{D}}{\operatorname{arg\,min}} \quad \underset{\mathclap{\substack{(\boldsymbol{X}, \boldsymbol{Y}) \sim \mathcal{D}}}}{\mathbb{E} \quad} [\mathcal{L}(f_{\boldsymbol{\theta}}(\boldsymbol{P}_{\mathcal{Z}}(\boldsymbol{X})), \boldsymbol{Y})]
\end{equation}
Therefore, the optimal prompt for task $\mathcal{Z}$ can be formulated as $\boldsymbol{P}_{\mathcal{Z}}^{*}(\boldsymbol{X})=(\boldsymbol{D}^{*}, \boldsymbol{E}^{\prime}, \boldsymbol{X}, \boldsymbol{V})$.

\subsection{Impact of Minor Lexical Changes}
\label{impact_of_minor_lexical}

The analysis presented in the previous section relies on a crucial premise: \ul{\emph{imperceptible lexical changes in prompts can significantly affect the model performance on downstream tasks}}. Before further explaining our approach, we first try to show the validity of this assumption.

Given an initially proven effective prompt, a model, and a predefined task pool, we attempt to demonstrate how prompts within the neighborhood of the original prompt influence the model's performance on the task pool. In this context, we broadly define a prompt's neighborhood as prompts that differ from the original by only \emph{one word} while maintaining a similar meaning. For instance, we consider "\emph{Does \ul{the} sentence make sense?}" to be within the neighborhood of "\emph{Does \ul{this} sentence make sense?}". To obtain these qualifying prompts, we employ a MLM. First, we iteratively replace each word in the task description with a \texttt{[MASK]} token. We then expect the MLM, by understanding the context, to provide the most probable fill-in words at each position based on the entire task description. After replacing the original words with the fill-in words at each position, we obtain a series of prompts within the neighborhood of the original prompt. We then evaluate the performance of each resulting prompt on the task pool.

\begin{figure*}[!t]
    \centering
  \includegraphics[width=1.0\textwidth]{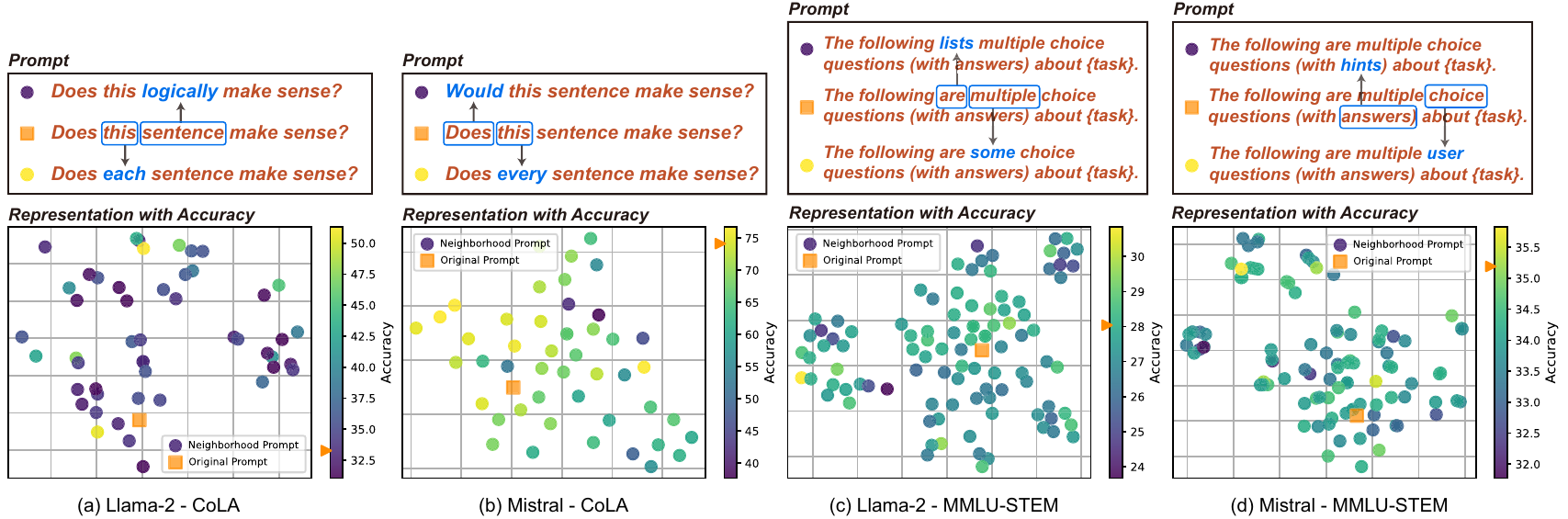}
  \caption{The visualization of model performance on CoLA and MMLU-STEM validation set with neighborhood prompts. The task description of the original prompt picked for CoLA is "\emph{Does this sentence make sense?}", and for MMLU-STEM is "\emph{The following are multiple choice questions (with answers) about \{task\}}", where \emph{\{task\}} is a placeholder to replace with detailed subset type, e.g., "abstract algebra". The point \begingroup\color[HTML]{3B3B3B}\ding{108}\endgroup \ in lighter color indicates better performance, and the square \begingroup\color[HTML]{FFA000}$\blacksquare$\endgroup \ indicates the original prompt, with the \begingroup\color[HTML]{FFA000}$\blacktriangleright$\endgroup \ in the color bar indicating the original performance. The \blueboxword{words} in the upper prompts indicate the changed words, and {\color[HTML]{006FEB}words} indicate the substitutions.}
  \label{word-change-acc}
\end{figure*}

Following these definitions and operations, we employ the validation sets of CoLA~\cite{Warstadt2019COLA} and MMLU-STEM~\cite{Hendrycks2021MMLU} subtasks, respectively, as predefined task pools. We use \texttt{Llama-2-7B-chat}~\cite{Touvron2023Llama2} and \texttt{Mistral-7B-Instruct-v0.1}~\cite{Jiang2023Mistral} as target models and use RoBERTa~\cite{Liu2019roberta} for obtaining neighborhood prompts. The initial prompts are picked from \emph{lm-evaluation-harness}~\cite{Gao2023evalharness}, and we generate ten most probable fill-in words for each position in task description. We then obtain their sentence representations in the target model and project them into a two-dimensional space using t-SNE~\cite{van2008visualizing}. Figure~\ref{word-change-acc} shows the performance of neighborhood prompts on downstream tasks with the distribution of their sentence representations. We can then reach several conclusions on the impact of minor lexical changes in prompts on downstream performance.

\ul{\emph{1.Semantically similar prompts have vastly different performances on downstream tasks, even if they differ by only one word.}} For example, when using neighborhood prompts on MMLU-STEM and \texttt{Llama-2-7B-chat} (Figure~\ref{word-change-acc}(c)), their performance differences can reach 7\%. Specifically, changing "\emph{The following \blueboxword{are} \blueboxword{multiple} choice questions (with answers) about \{task\}.}" (\begingroup\color[HTML]{FFA000}$\blacksquare$\endgroup) to "\emph{The following {\color[HTML]{006FEB}lists} multiple choice questions (with answers) about \{task\}.}" (\begingroup\color[HTML]{6A0EA1}\ding{108}\endgroup) reduces the accuracy from the original 28.04\% to 23.92\%, while changing it to "\emph{The following are {\color[HTML]{006FEB}some} choice questions (with answers) about \{task\}.}" (\begingroup\color[HTML]{F3F800}\ding{108}\endgroup) increases the accuracy to 30.86\%. Intuitively, such minor lexical variations should have a minimal impact on semantics, and human performance would likely remain consistent when completing downstream tasks guided by these three prompts~\cite{drewnowski1978universal,mccusker1981word,shillcock1981eye,rayner2006raeding}. However, models exhibit a high degree of sensitivity to these changes.

\ul{\emph{2.In the latent representation space, prompts that are in close proximity may have vastly different performance on downstream tasks.}} In most cases, the performance of neighborhood prompts on downstream tasks does not demonstrate a clear correlation with the distribution of their sentence representations. Even when the representations of prompts are clustered together, they can still have substantial performance discrepancies. For example, in the representation space of \texttt{Llama-2-7B-chat}, the best-performing prompt (51.2\%, \begingroup\color[HTML]{F3F800}\ding{108}\endgroup) on CoLA (Figure~\ref{word-change-acc}(a)) is situated in very close proximity to the prompt with nearly the worst performance (32.4\%, \begingroup\color[HTML]{6A0EA1}\ding{108}\endgroup). From the perspective of sentence representations, this observation indicates that even for semantically highly similar prompts, their performance may be vastly different, and it is difficult to infer their performance from one another directly.

\subsection{COPLE}
\label{cople}

According to our findings, even for semantically similar prompts with only one word difference, their performance on downstream tasks may be very different, and we cannot infer the performance of one prompt from another seemingly similar prompt. Therefore, we propose COPLE, trying to recover the 
degraded ability of models caused by lexical sensitivity. The key idea behind COPLE is to guide the lexical optimization of the initial prompt by the model performance on a batch of reference tasks i.i.d. to the downstream tasks, and iteratively improve the prompt based on the feedback from these references to converge towards an optimal prompt that maximizes performance across the task distribution. Specifically, COPLE consists of the following four parts:

\paragraph{Proxy Reference Tasks.}

To find the optimal $\boldsymbol{D}^{*}$ defined in (\ref{combinatorial_optimization}), while avoiding data leakage, we first randomly sample a batch of reference tasks from the same distribution $\mathcal{D}$, denoted as $\mathcal{Z}_{\textit{ref}}$. For example, when targeting the validation set of a dataset as downstream tasks, we construct the reference tasks by sampling from the training set. These reference tasks serve as a proxy for evaluating the prompt on the task distribution, which also accelerates COPLE, as evaluating on a small batch of examples is not as expensive as evaluating on the full validation set. Therefore, the optimal task description that COPLE tries to find can be transformed from (\ref{combinatorial_optimization}) to:
\begin{equation}
\begin{aligned}
\label{reference_optimization}
        \boldsymbol{D}^{*} &= \underset{\boldsymbol{D}}{\operatorname{arg\,min}} \quad \mathcal{L}_{\textit{ref}}(\boldsymbol{D}) \\
        &= \underset{\boldsymbol{D}}{\operatorname{arg\,min}} \quad \underset{\mathclap{\substack{(\boldsymbol{X}, \boldsymbol{Y}) \sim \mathcal{Z}_{\textit{ref}}}}}{\mathbb{E} \quad} [\mathcal{L}(f_{\boldsymbol{\theta}}(\boldsymbol{P}_{\mathcal{Z}_{\textit{ref}}}(\boldsymbol{X})), \boldsymbol{Y})]
\end{aligned}
\end{equation}
where $\boldsymbol{P}_{\mathcal{Z}_{\textit{ref}}}$ denotes the entire prompt for the reference tasks $\mathcal{Z}_{\textit{ref}}$.

\paragraph{Search by Word Influence.}
With the proxy reference tasks, COPLE then performs an iterative optimization process to find the optimal task description $\boldsymbol{D}^*$. As COPLE serves as a black-box method without accessing the gradient information of the model, we first define the influence of each word in the task description as the expected performance difference on proxy tasks when the word is deleted from the task description. Formally:
\begin{equation}
\label{influence}
    \mathcal{I}(d_i) = |\mathcal{L}_{\textit{ref}}(\boldsymbol{D}) - \mathcal{L}_{\textit{ref}}(\boldsymbol{D}_{\backslash i})|
\end{equation}
where $\boldsymbol{D}_{\backslash i}$ denotes the task description with token $d_i$ removed. For efficiency purposes, COPLE obtains the influence of each word only on the initial task description. Then, COPLE tries to iteratively find the optimal substitution for the most influential words in the descending order of their influence.

\paragraph{Lexical Search Space.}
To construct the search space $\mathcal{C}_i$ for token $d_i$ in the task description, similar to that in \S\ref{impact_of_minor_lexical}, we reuse a pre-trained MLM to find semantically similar words. Formally, at each iteration $t$, we mask out the target $d_i$ in current task description and feed the masked description into a pre-trained MLM $f_{\textit{MLM}}$. The MLM then predicts a probability distribution over its vocabulary $\mathcal{V}$ for the masked position:
\begin{equation}
\label{mlm-search-space}
    p(w|\boldsymbol{D}_{\backslash i}^{(t)}) = f_{\textit{MLM}}(d_1, \ldots, \texttt{[MASK]}, \ldots, d_n)
\end{equation}
where $w \in \mathcal{V}$, $\boldsymbol{D}_{\backslash i}^{(t)}$ denotes the task description at iteration $t$ with token $d_i$ masked out. We then select the top-$k$ words with the highest probabilities and a empty token (delete) as the candidates.

\paragraph{Iterative Optimization.} 
At each iteration $t$, COPLE selects the most influential token that has not been searched and constructs its corresponding search space according to (\ref{mlm-search-space}). For each candidate $c \in \mathcal{C}_{i}$, COPLE substitutes $d_{i}$ with $c$ and evaluates the performance of the updated task description $\boldsymbol{D}^{(t)}$ on the small proxy reference tasks:
\begin{equation}
\label{update_description}
    \mathcal{L}_{\textit{ref}}(\boldsymbol{D}^{(t)}) = \underset{\mathclap{\substack{(\boldsymbol{X}, \boldsymbol{Y}) \sim \mathcal{Z}_{\textit{ref}}}}}{\mathbb{E} \quad} [\mathcal{L}(f_{\boldsymbol{\theta}}(\boldsymbol{P}_{\mathcal{Z}_{\textit{ref}}}^{(t)}(\boldsymbol{X})), \boldsymbol{Y})]
\end{equation}
where $\boldsymbol{D}^{(t)} = (d_1, \ldots, c, \ldots, d_n)$, and $\smash{\boldsymbol{P}_{\mathcal{Z}_{\textit{ref}}}^{(t)}}$ is the prompt with the task description $\boldsymbol{D}^{(t)}$ at iteration $t$. COPLE then selects the candidate $c^*$ that minimizes the expected loss on the proxy reference tasks:
\begin{equation}
    c^* = \underset{c \in \mathcal{C}_{i}}{\operatorname{arg\,min}} \quad \mathcal{L}_{\textit{ref}}(\boldsymbol{D}^{(t)})
\end{equation}
and updates the task description to $\smash{\boldsymbol{D}^{(t+1)}}$ by replacing $d_{i}$ with $c^*$ if its performance is better than $\boldsymbol{D}^{(t)}$, i.e., if $\smash{\mathcal{L}_{\textit{ref}}(\boldsymbol{D}^{(t+1)}) < \mathcal{L}_{\textit{ref}}(\boldsymbol{D}^{(t)})}$:
\begin{equation}
    \boldsymbol{D}^{(t+1)} = (d_1, \ldots, c^*, \ldots, d_n)
\end{equation}
otherwise, $\smash{\boldsymbol{D}^{(t+1)}}$ is kept the same as $\smash{\boldsymbol{D}^{(t)}}$.
This process is repeated until all the most influential words are traversed (detailed in \S\ref{setup}).
Ideally, we take the found best $\boldsymbol{D}$ after optimization as the $\boldsymbol{D}^{*}$ defined in (\ref{reference_optimization}). The final optimized task description $\boldsymbol{D}^*$ is then used to construct the optimal prompt $\boldsymbol{P}_{\mathcal{Z}}^{*}(\boldsymbol{X})$ for the downstream tasks.

\section{Experiment}

\subsection{Experiment Setup}
\label{setup}

\begin{table*}[tb]
\centering
\sisetup{
  table-format=2.2,       %
  table-space-text-pre={\textsuperscript{***}},  %
  table-space-text-post={\textsubscript{***}},   %
  table-align-text-post=false,                   %
}
\resizebox{1.0\textwidth}{!}{%
\begin{tabular}{l
S[table-format=2.2(1)]
S[table-format=2.2(1)]
S[table-format=2.2(1)]
S[table-format=2.2(1)]
S[table-format=2.2(1)]
S[table-format=2.2(1)]
S[table-format=2.2(1)]
S[table-format=2.2(1)]
S[table-format=2.2(1)]
S[table-format=2.2(1)]
S[table-format=2.2(1)]}
\toprule
                               & \multicolumn{7}{c|}{GLUE} & \multicolumn{4}{c}{MMLU} \\ \midrule
                         & {\hspace{-5mm}SST2}         & {\hspace{-5mm}CoLA}          & {\hspace{-5mm}MNLI}           & {\hspace{-5mm}QNLI}           & {\hspace{-5mm}RTE}         & {\hspace{-5mm}MRPC}  & {\hspace{-5mm}QQP}  & {\hspace{-5mm}STEM}  & {\hspace{-5mm}Humanities} & {\hspace{-4mm}Soc.Sci} & {\hspace{-5mm}Other}  \\ \midrule \midrule
\multicolumn{12}{c}{\texttt{Llama-2-7B-chat}}                                                                                        \\ \midrule

Original                     & 90.71 & 33.27 & 35.51 & 51.57 & 53.07 & 68.06 & 27.58 & 28.04 & 23.94 & 35.31 & 38.59       \\
{\hspace{2mm} \emph{\ \ w/ }COPLE} & \, \textbf{92.43}\textsubscript{0.70} & \, \textbf{65.72}\textsubscript{1.29} & \, \textbf{52.42}\textsubscript{1.33} & \, \textbf{69.38}\textsubscript{2.10} & \, \textbf{68.59}\textsubscript{4.08} & \, \textbf{68.17}\textsubscript{0.53} & \, \textbf{57.11}\textsubscript{1.80} & \, \textbf{31.46}\textsubscript{0.54} & \, \textbf{27.90}\textsubscript{1.77} & \, \textbf{37.09}\textsubscript{0.79} & \, \textbf{46.20}\textsubscript{0.40}    \\ \midrule
\multicolumn{12}{c}{\texttt{Mistral-7B-Instruct-v0.1}}                                                                                        \\ \midrule
Original                     & 87.27 & 74.31 & 65.18 & 75.22 & 64.98 & 54.41 & 68.92 & 35.19 & 37.07 & 52.23 & 51.83 \\ 
{\hspace{2mm} \emph{\ \ w/ }COPLE} & \, \textbf{91.21}\textsubscript{1.48} & \, \textbf{78.24}\textsubscript{1.58} & \, \textbf{65.37}\textsubscript{0.89} & \, \textbf{79.34}\textsubscript{0.35} & \, \textbf{71.60}\textsubscript{0.55} & \, \textbf{71.08}\textsubscript{0.98} & \, \textbf{75.93}\textsubscript{0.15} & \, \textbf{35.83}\textsubscript{0.54} & \, \textbf{37.45}\textsubscript{0.51} & \, \textbf{53.02}\textsubscript{0.17} & \, \textbf{52.96}\textsubscript{0.28} \\ \midrule
\multicolumn{12}{c}{ChatGPT (\texttt{gpt-3.5-turbo-0125})}                                                                                        \\ \midrule
Original & 94.38 & 80.73 &  $\backslash$  &  $\backslash$  & 62.09 & 41.91 &  $\backslash$  & 34.16 & 42.47 & 58.16 & 57.46 \\
{\hspace{2mm} \emph{\ \ w/ }COPLE} & \, \textbf{94.80}\textsubscript{0.17} & \, \textbf{82.91}\textsubscript{0.16} & \, $\backslash$ & $\backslash$ & \, \textbf{80.35}\textsubscript{0.34} & \, \textbf{70.75}\textsubscript{0.37} & \, $\backslash$ & \, \textbf{36.45}\textsubscript{0.76} & \, \textbf{43.44}\textsubscript{0.39} & \, \textbf{58.46}\textsubscript{0.59} & \, \textbf{57.61}\textsubscript{0.14} \\ \bottomrule

\end{tabular}%
}
\caption{Performance comparison (Accuracy) of models on GLUE and MMLU benchmarks using the human-crafted prompts (\emph{Original}) with and without applying COPLE. The \textbf{bold} values indicate the better results, while the standard deviations are provided in smaller font. For MNLI, we report the average results on the \emph{matched} and \emph{mismatched} subsets.  Some results for \texttt{gpt-3.5-turbo-0125} are denoted as "$\backslash$", indicating that, due to the huge validation set and cost and efficiency considerations, corresponding experiments are not conducted.}
\label{main-acc-ori}
\vspace{-2mm}
\end{table*}

\paragraph{Dataset and Model.}
We use GLUE~\cite{Wang2019GLUE} and MMLU~\cite{Hendrycks2021MMLU} for evaluation. For GLUE, we report the results on SST2~\cite{Socher2013SST2}, CoLA~\cite{Warstadt2019COLA}, MNLI~\cite{Williams2018MNLI}, QNLI~\cite{Rajpurkar2016QNLI}, RTE~\cite{Giampiccolo2007RTE}, MRPC~\cite{Dolan2005MRPC}, and QQP~\cite{Cer2017QQP}. For MMLU, we separately report the results on the subset of STEM, Humanities, Social Sciences, and Other.
We use the \texttt{Llama-2-7B-chat}~\cite{Touvron2023Llama2}, \texttt{Mistral-7B-Instruct-v0.1}~\cite{Jiang2023Mistral}, and ChatGPT (\texttt{gpt-3.5-turbo-0125})~\cite{Openai2022ChatGPT} as the target model. 
Please see Appendix~\ref{appendix_dataset} for more details on datasets and models.

\paragraph{Baseline.}

To show the effectiveness of COPLE and empirically demonstrate the conclusions in \S\ref{impact_of_minor_lexical}, we evaluate COPLE in the following scenarios: (i) \emph{Original}: using human-crafted prompts from \emph{HELM}~\cite{Lee2023HELM} and \emph{lm-evaluation-harness}~\cite{Gao2023evalharness}. (ii) \emph{In-context Learning}, following \citet{Brown2020GPT3}, randomly concatenating 1 and 3 examples from the training set with \emph{Original} manual prompts (as the $\boldsymbol{E}^{\prime}$ in $\boldsymbol{P}_{\mathcal{Z}}(\boldsymbol{X})$), denoted as the \emph{1-shot} and \emph{3-shot} settings, respectively. (iii) \emph{Emotion Prompt}: combining two different self-monitoring style emotional stimuli, used in \cite{Li2023EmotionPrompt} with \emph{Original} manual prompts, denoted as \emph{EP02} and \emph{EP03}, respectively. (iv) \emph{Chain-of-thought}: combining zero-shot CoT trigger~\cite{KojimaGRMI22} with \emph{Original} manual prompts, denoted as \emph{Zero-shot-CoT}. Please see Appendix \ref{appendix_prompt} for the detailed prompts used in evaluation.

\paragraph{Implementation Details.}

To construct the proxy reference tasks $\mathcal{Z}_{\textit{ref}}$, we sample 100 tasks from training set.
For the search space, we use RoBERTa~\cite{Liu2019roberta} as the MLM in (\ref{mlm-search-space}), selecting the top-30 highest probability substitutions for each iteration. Following (\ref{influence}), we take the 70\% most influential words in a task description to perform optimization. We use HELM-style evaluation, with more details available in Appendix \ref{appendix_evaluation}. All reported average results and standard deviations are obtained from 3 runs with different seeds.

\subsection{Main Results}

\paragraph{Popular Prompts Suffer From Lexical Sensitivity.} 

Table~\ref{main-acc-ori} shows the model performance on different tasks using \emph{Original} human-crafted prompts and related prompts optimized by COPLE. The results demonstrate that even widely used human-crafted prompts fail to maximize model performance on downstream tasks, due to lexical sensitivity and specific words in prompts. Specifically, for \texttt{Llama-2-7B-chat}, the average accuracy across all datasets increased from 44.15\% to 56.04\% (11.89\%$\uparrow$) after applying COPLE. For \texttt{Mistral-7B-Instruct-v0.1}, the average accuracy increased from 60.60\% to 64.73\% (4.13\%$\uparrow$). ChatGPT also exhibited a notable improvement, with the average accuracy increasing from 58.92\% to 65.59\% (6.67\%$\uparrow$, 4 GLUE datasets $+$ MMLU) when using prompts optimized by COPLE.

\begin{table*}[tb]
\centering
\sisetup{
  table-format=2.2,       %
  table-space-text-pre={\textsuperscript{***}},  %
  table-space-text-post={\textsubscript{***}},   %
  table-align-text-post=false,                   %
}
\resizebox{1.0\textwidth}{!}{%
\begin{tabular}{l
S[table-format=2.2(1)]
S[table-format=2.2(1)]
S[table-format=2.2(1)]
S[table-format=2.2(1)]
S[table-format=2.2(1)]
S[table-format=2.2(1)]
S[table-format=2.2(1)]
S[table-format=2.2(1)]
S[table-format=2.2(1)]
S[table-format=2.2(1)]
S[table-format=2.2(1)]}
\toprule
                               & \multicolumn{7}{c|}{GLUE} & \multicolumn{4}{c}{MMLU} \\ \midrule
                         & {\hspace{-5mm}SST2}         & {\hspace{-5mm}CoLA}          & {\hspace{-5mm}MNLI}           & {\hspace{-5mm}QNLI}           & {\hspace{-5mm}RTE}         & {\hspace{-5mm}MRPC}  & {\hspace{-5mm}QQP}  & {\hspace{-5mm}STEM}  & {\hspace{-5mm}Humanities} & {\hspace{-4mm}Soc.Sci} & {\hspace{-5mm}Other}  \\ \midrule \midrule
\multicolumn{12}{c}{\texttt{Llama-2-7B-chat}}                                                                                        \\ \midrule
\multicolumn{12}{l}{\emph{In-Context Learning Prompts}} \\
1-shot & 57.68 & 68.74 & 19.81 & 49.57 & 51.62 & 68.14 & 3.48 & 26.17 & 21.43 & 25.82 & 22.98 \\
{\hspace{2mm} \emph{\ \ w/ }COPLE} & \, \textbf{88.13}\textsubscript{3.16} & \, \textbf{69.13}\textsubscript{0.29} & \, \textbf{38.46}\textsubscript{5.36} & \, \textbf{52.61}\textsubscript{0.62} & \, \textbf{55.23}\textsubscript{0.51} & \, \textbf{68.23}\textsubscript{0.12} & \, \hspace{1mm}\textbf{45.93}\textsubscript{17.25} & \, \textbf{28.97}\textsubscript{0.31} & \, \textbf{24.15}\textsubscript{0.27} & \, \textbf{30.12}\textsubscript{0.63} & \, \textbf{23.38}\textsubscript{0.23} \\
3-shot & 51.61 & 67.98 & 0.03 & 51.91 & 56.51 & 68.38 & 23.03 & 26.48 & 24.13 & 27.00 & 22.82 \\
{\hspace{2mm} \emph{\ \ w/ }COPLE} & \, \textbf{70.07}\textsubscript{1.62} & \, \textbf{68.36}\textsubscript{0.38} & \, \textbf{31.17}\textsubscript{0.19} & \, \textbf{55.78}\textsubscript{0.22} & \, \textbf{57.76}\textsubscript{0.26} & \, \textbf{68.59}\textsubscript{0.28} & \, \textbf{57.61}\textsubscript{1.22} & \, \textbf{30.22}\textsubscript{0.88} & \, \textbf{26.38}\textsubscript{0.11} & \, \textbf{29.82}\textsubscript{0.21} & \, \textbf{25.45}\textsubscript{0.33} \\ \midrule
\multicolumn{12}{l}{\emph{Emotion Prompts}} \\
EP02 & 85.32 & 31.16 & 35.09 & 53.25 & 53.07 & 67.92 & 8.58 & 26.17 & 21.24 & 30.27 & 36.89 \\
{\hspace{2mm} \emph{\ \ w/ }COPLE} & \, \textbf{92.26}\textsubscript{0.57} & \, \textbf{67.98}\textsubscript{0.54} & \, \textbf{50.15}\textsubscript{3.93} & \, \textbf{64.59}\textsubscript{1.85} & \, \textbf{57.16}\textsubscript{4.59} & \, \textbf{68.42}\textsubscript{0.34} & \, \textbf{30.55}\textsubscript{0.60} & \, \textbf{31.78}\textsubscript{0.44} & \, \textbf{28.09}\textsubscript{2.59} & \, \textbf{35.91}\textsubscript{1.68} & \, \textbf{40.14}\textsubscript{0.20} \\
EP03 & 91.51 & 36.53 & 35.66 & 50.36 & 54.15 & 68.38 & 17.18 & 25.23 & 21.62 & 34.42 & 39.15 \\
{\hspace{2mm} \emph{\ \ w/ }COPLE} & \, \textbf{92.78}\textsubscript{0.65} & \, \textbf{68.41}\textsubscript{1.02} & \, \textbf{52.23}\textsubscript{0.03} & \, \textbf{67.40}\textsubscript{3.60} & \, \textbf{62.09}\textsubscript{3.06} & \, \textbf{68.63}\textsubscript{0.35} & \, \textbf{42.34}\textsubscript{6.23} & \, \textbf{33.02}\textsubscript{1.32} & \, \textbf{27.12}\textsubscript{0.14} & \, \textbf{36.65}\textsubscript{2.31} & \, \textbf{43.80}\textsubscript{0.20} \\ \midrule
\multicolumn{12}{l}{\emph{Chain-of-thought Prompts}} \\
Zero-shot CoT & 68.23 & 65.19 & 41.51 & 62.97 & 55.96 & 50.74 & 11.63 & 25.55 & 30.12 & 27.89 & 24.23 \\
{\hspace{2mm} \emph{\ \ w/ }COPLE} & \, \textbf{84.29}\textsubscript{3.92} & \, \textbf{67.45}\textsubscript{1.02} & \, \textbf{50.51}\textsubscript{1.03} & \, \textbf{69.64}\textsubscript{0.92} & \, \textbf{59.81}\textsubscript{0.83} & \, \textbf{66.42}\textsubscript{1.04} & \, \textbf{23.49}\textsubscript{0.88} & \, \textbf{27.73}\textsubscript{0.54} & \, \textbf{32.72}\textsubscript{0.68} & \, \textbf{34.42}\textsubscript{1.26} & \, \textbf{31.27}\textsubscript{1.59} \\ \midrule \midrule
\multicolumn{12}{c}{\texttt{Mistral-7B-Instruct-v0.1}}                                                                                        \\ \midrule
\multicolumn{12}{l}{\emph{In-Context Learning Prompts}} \\
1-shot & 81.08 & 34.13 & 58.87 & 51.55 & 51.99 & 31.86 & 43.64 & 28.66 & 31.66 & 42.14 & 45.92 \\
{\hspace{2mm} \emph{\ \ w/ }COPLE} & \, \textbf{91.80}\textsubscript{0.41} & \, \textbf{72.42}\textsubscript{0.77} & \, \textbf{60.72}\textsubscript{1.46} & \, \textbf{71.94}\textsubscript{0.18} & \, \textbf{65.16}\textsubscript{2.81} & \, \textbf{65.07}\textsubscript{0.52} & \, \textbf{64.96}\textsubscript{0.16} & \, \textbf{31.15}\textsubscript{0.44} & \, \textbf{31.98}\textsubscript{0.73} & \, \textbf{44.31}\textsubscript{0.95} & \, \textbf{46.90}\textsubscript{0.28} \\
3-shot & 91.28 & 66.73 & 41.45 & 69.51 & 53.07 & 56.13 & 73.39 & 31.15 & 39.58 & 45.99 & 50.70 \\
{\hspace{2mm} \emph{\ \ w/ }COPLE} & \, \textbf{93.43}\textsubscript{0.07} & \, \textbf{72.77}\textsubscript{0.27} & \, \textbf{56.35}\textsubscript{0.68} & \, \textbf{75.62}\textsubscript{0.58} & \, \textbf{68.77}\textsubscript{5.87} & \, \textbf{67.65}\textsubscript{0.35} & \, \textbf{75.42}\textsubscript{0.70} & \, \textbf{35.31}\textsubscript{2.12} & \, \textbf{39.67}\textsubscript{0.14} & \, \textbf{49.11}\textsubscript{1.05} & \, \textbf{51.97}\textsubscript{0.60} \\ \midrule
\multicolumn{12}{l}{\emph{Emotion Prompts}} \\
EP02 & 64.56 & 70.18 & 62.83 & 62.77 & 53.07 & 38.24 & 61.75 & 35.23 & 35.71 & 50.15 & 49.86 \\
{\hspace{2mm} \emph{\ \ w/ }COPLE} & \, \textbf{89.11}\textsubscript{1.78} & \, \textbf{75.36}\textsubscript{0.41} & \, \textbf{64.36}\textsubscript{0.82} & \, \textbf{72.57}\textsubscript{1.54} & \, \textbf{72.56}\textsubscript{2.04} & \, \textbf{70.10}\textsubscript{0.69} & \, \textbf{72.43}\textsubscript{0.98} & \, \textbf{37.07}\textsubscript{0.44} & \, \textbf{37.64}\textsubscript{0.82} & \, \textbf{52.08}\textsubscript{0.30} & \, \textbf{51.97}\textsubscript{1.00} \\
EP03 & 76.61 & 74.49 & 65.56 & 74.24 & 64.62 & 62.75 & 73.37 & 34.58 & 36.87 & 51.93 & 50.99 \\
{\hspace{2mm} \emph{\ \ w/ }COPLE} & \, \textbf{91.44}\textsubscript{0.33} & \, \textbf{76.32}\textsubscript{0.14} & \, \textbf{67.24}\textsubscript{0.36} & \, \textbf{76.37}\textsubscript{0.91} & \, \textbf{72.38}\textsubscript{1.28} & \, \textbf{74.39}\textsubscript{0.17} & \, \textbf{75.10}\textsubscript{0.90} & \, \textbf{36.76}\textsubscript{0.88} & \, \textbf{37.45}\textsubscript{0.27} & \, \textbf{52.67}\textsubscript{0.21} & \, \textbf{52.02}\textsubscript{0.16} \\ \midrule
\multicolumn{12}{l}{\emph{Chain-of-thought Prompts}} \\
Zero-shot CoT & 86.01 & 76.03 & 50.04 & 78.29 & 66.06 & 65.21 & 72.74 & 35.19 & 35.91 & 50.74 & 50.14 \\
{\hspace{2mm} \emph{\ \ w/ }COPLE} & \, \textbf{90.90}\textsubscript{0.26} & \, \textbf{76.27}\textsubscript{1.02} & \, \textbf{67.41}\textsubscript{0.68} & \, \textbf{79.45}\textsubscript{0.45} & \, \textbf{71.48}\textsubscript{0.63} & \, \textbf{74.14}\textsubscript{0.52} & \, \textbf{76.80}\textsubscript{3.13} & \, \textbf{36.34}\textsubscript{0.18} & \, \textbf{36.87}\textsubscript{0.33} & \, \textbf{52.08}\textsubscript{0.21} & \, \textbf{52.54}\textsubscript{0.20} \\ \bottomrule

\end{tabular}%
}
\caption{Performance comparison (Accuracy) of models on GLUE and MMLU using different initial prompts with and without applying COPLE. The \textbf{bold} and smaller values denote better results and standard deviations.}
\label{main-acc-more}
\vspace{-2mm}
\end{table*}

\paragraph{COPLE Recovers the Ability on Both Instruct-Following and Solving Downstream Tasks.} 

Table~\ref{main-acc-more} shows the model performance on various tasks using different prompts. Recall that the HELM-style evaluation used in our experiments may yield performance worse than random guessing, suggesting that models fail to complete the task as instructed by the provided prompt. Following this, we find that further modifications to the \emph{Original} prompts, such as adding few-shot demo examples, may not always improve the performance. Such modifications may lead to the generation of incorrect or entirely irrelevant responses, such as repeating demo examples or generating non-existent new examples. Therefore, it can be deduced that the decline in model performance on downstream tasks may be attributed to a decreased ability of (i) problem-solving, as when the model gives wrong results, and (ii) instruct-following, as when the model gives irrelevant results. For example, for \texttt{Llama-2-7B-chat} on QQP, the \emph{3-shot} accuracy decreases from 27.58\% to 23.03\% (4.55\%$\downarrow$) compared to \emph{Original}. However, without complex prompt engineering, minor lexical optimization performed by COPLE is enough to recover the declined ability, as the accuracy increases from 23.03\% to 57.61\% after applying COPLE, which also outperforms the original prompt optimized by COPLE (Table~\ref{main-acc-ori}, 57.11\%). 
Please see Appendix \ref{appendix_optimized_prompt} for the detailed optimized prompts.

\subsection{Analysis and Ablation Study}

In this section, we conduct further analysis and ablation studies on COPLE. When not specified, the results are obtained on \texttt{Llama-2-7B-chat}.

\paragraph{Difference Between Prompts.}

\begin{table}[tb]
\centering
\resizebox{1.0\columnwidth}{!}{%
\begin{tabular}{c|
c|
S[table-format=3.0]|
S[table-format=1.2]
S[table-format=1.2]
S[table-format=3.0]|
S[table-format=1.2]
S[table-format=1.2]
S[table-format=3.0]
}
\toprule
                          &  &   & \multicolumn{3}{c|}{\texttt{Llama-2-7B-chat}} & \multicolumn{3}{c}{\texttt{Mistral-7B-Instruct-v0.1}} \\ \cline{4-9}
                    &  &    {PPL.ori}    & {U.Sim}  & {BERTScore}  & {PPL}  & {U.Sim}  & {BERTScore}  & {PPL}  \\ \midrule \midrule
\multirow{7}{*}{GLUE}       &  SST2  & 46                & 0.84 & 0.89 & 163 & 0.85 & 0.92 & 82      \\
                           &CoLA  &  67                 & 0.79 & 0.75 & 1875 & 0.81 & 0.87 & 412      \\ 
                           &MNLI  & 635    & 0.75 & 0.82 & 805 & 0.78 & 0.88 & 1781 \\
                           & QNLI  & 206  & 0.77 & 0.82 & 487 & 0.78 & 0.85 & 521 \\
                           & RTE  & 635  & 0.74 & 0.78 & 2026 &  0.78 & 0.83 & 842 \\
                           & MRPC  & 58  & 0.78 & 0.87 & 103 & 0.78 & 0.87 & 326 \\
                           & QQP & 185  & 0.72  & 0.76 & 656 & 0.78 & 0.86 & 306 \\ \midrule
\multirow{4}{*}{MMLU}       & STEM  &   {\multirow{4}{*}{276}}             & 0.76 & 0.88 & 515 & 0.79 & 0.88 & 408  \\ 
                           & Humanities  &   & 0.79 & 0.89 & 529 & 0.79 & 0.92 & 491 \\ 
                           & Soc.Sci  &   & 0.80 & 0.91 & 494 & 0.80 & 0.94 & 479 \\
                           & Other  &   & 0.78 & 0.84 & 663 & 0.79 & 0.87 & 525 \\ \midrule \midrule
                           & Avg.  & 267 & 0.77  & 0.84  & 756 & 0.79 & 0.88  & 561 \\

                           \bottomrule 

\end{tabular}%
}
\caption{Difference of semantic similarity and perplexity between original prompts and optimized prompts. \emph{U.Sim} denotes the similarity obtained through USE.}
\label{main-quality}
\vspace{-3mm}
\end{table}

To measure the difference between original prompts and optimized prompts, we utilize Universal Sentence Encoder (USE)~\cite{cer2018universal} and BERTScore~\cite{ZhangKWWA20} to obtain semantic similarity. 
We also obtain their perplexity (PPL)~\cite{jelinek1977perplexity} with GPT-2~\cite{Radford2018GPT2}. Table~\ref{main-quality} illustrates the differences between prompts. The USE similarity and BERTScore between the original and optimized prompts are consistently high across all tasks, indicating that the semantics of the prompts are well-preserved after optimization, which also confirms our conclusions in \S\ref{impact_of_minor_lexical}. However, the perplexity of the optimized prompts increases significantly compared to the original, indicating that using challenging words for the language model in prompts, rather than common words picked by humans, may help to improve the model performance on solving downstream tasks.

\paragraph{$\#$.Word Change in Prompt.}

\begin{figure}[!tbp]
    \centering
  \includegraphics[width=1.0\columnwidth]{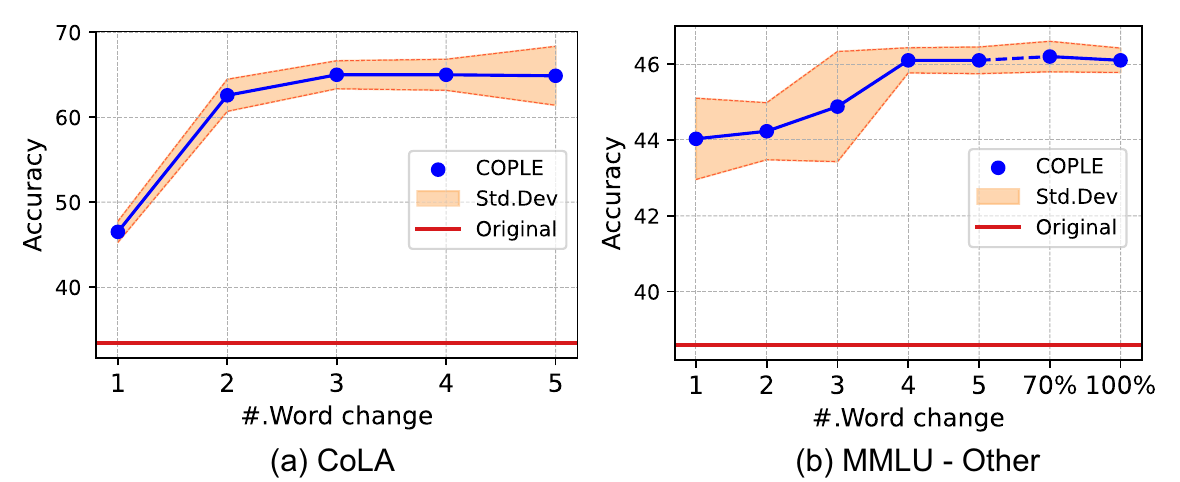}
  \caption{Impact of the number of words changed in prompt on downstream performance.}
  \label{ablation_word-rate}
  \vspace{-3mm}
\end{figure}

Figure~\ref{ablation_word-rate} shows the impact of the number of words changed in prompt.
Even if only changing the one word with the highest influence, the model performance significantly improves (CoLA: 33.27\% to 46.50\%$_{\pm \text{1.23\%}}$, MMLU-Other: 38.59\% to 44.04\%$_{\pm \text{1.07\%}}$). When the \emph{$\#$.Word Change} increases, COPLE tends to achieve higher accuracy, while a moderate value is enough to achieve nearly the optimal result.

\begin{figure}[!tbp]
    \centering
  \includegraphics[width=1.0\columnwidth]{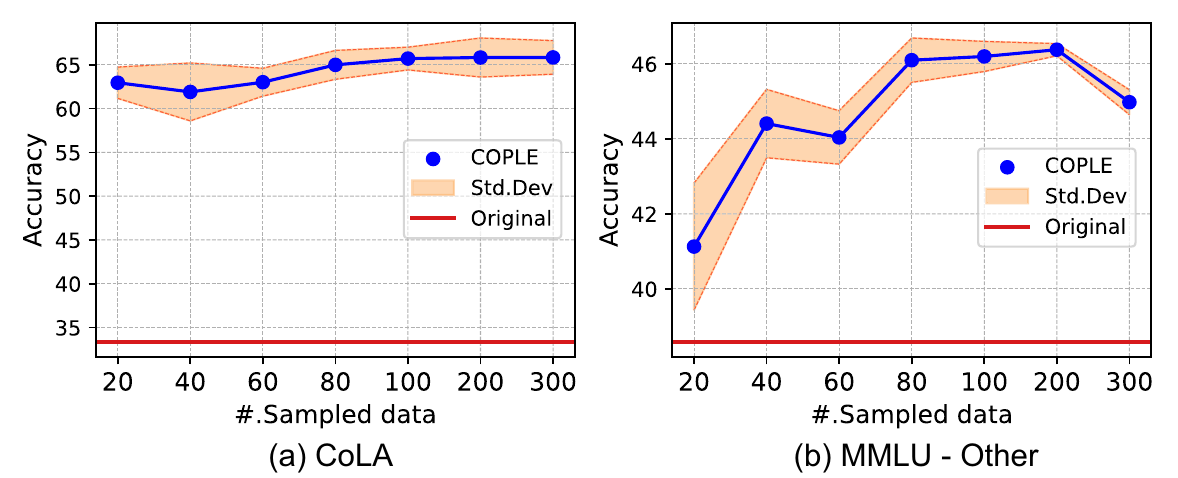}
  \caption{Impact of the number of sampled examples in proxy reference tasks on downstream performance.}
  \label{ablation_sample-rate}
  \vspace{-3mm}
\end{figure}
\begin{figure}[!tbp]
    \centering
  \includegraphics[width=1.0\columnwidth]{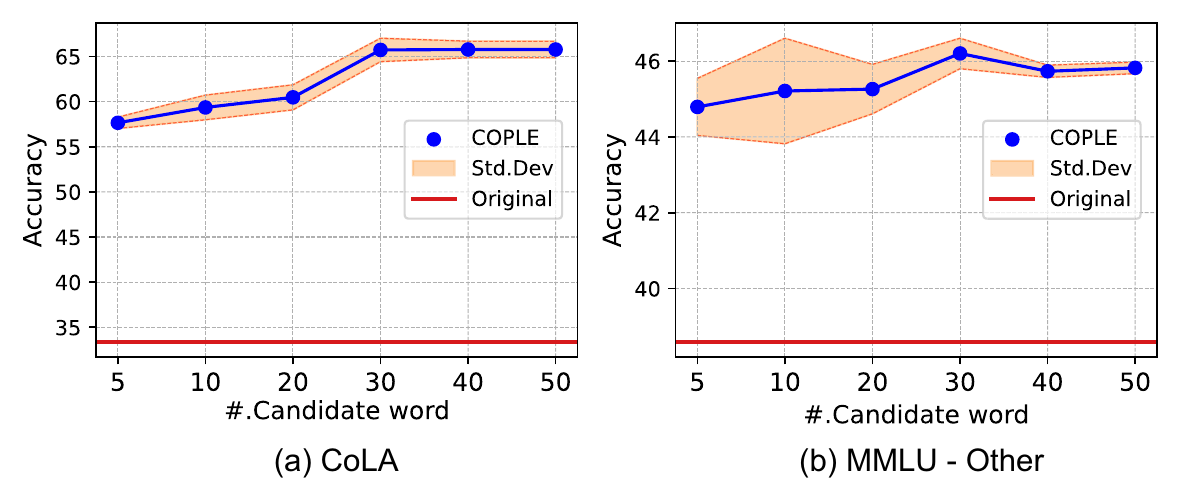}
  \caption{Impact of the number of candidate words in search space on downstream performance.}
  \label{ablation_max-candidate}
  \vspace{-2mm}
\end{figure}

\paragraph{$\#$.Sampled Data for Proxy Tasks ($|\mathcal{Z}_{\textit{ref}}|$).}

Figure~\ref{ablation_sample-rate} shows the impact of the size of proxy tasks.
When proxy tasks contain a small number of 20 examples, COPLE still achieves notable improvements (CoLA: 33.27\% to 62.96\%$_{\pm \text{1.78\%}}$; MMLU-Other: 38.59\% to 41.13\%$_{\pm \text{1.69\%}}$). However, a larger size of sampled data helps find prompts with higher accuracy and lower standard deviations.

\paragraph{$\#$.Candidate Word in Search Space ($k$).}

Figure~\ref{ablation_max-candidate} shows the impact of the number of candidate words in search space. 
A small $k$ is enough to support COPLE achieving considerable improvement ($k$=5, CoLA: 33.27\% to 57.65$_{\pm \text{0.62\%}}$, MMLU-Other: 38.59\% to 44.79$_{\pm \text{0.75\%}}$). Therefore, for efficient purposes, a small $k$ is more suitable. However, using larger search space for each word is more likely to find prompts with better performance.

\paragraph{Word Influence.}

Table~\ref{main-ablation-random} shows the ablation results of the search strategy related to word influence in COPLE. When replacing the search strategy with the random method, the average performance optimized by COPLE on MMLU decreases from 35.66\% to 34.17\%, and COPLE is less stable as the standard deviation gets larger.

\begin{table}[!tb]
\centering
\sisetup{
  table-format=2.2,       %
  table-space-text-pre={\textsuperscript{***}},  %
  table-space-text-post={\textsubscript{***}},   %
  table-align-text-post=false,                   %
}
\resizebox{1.0\columnwidth}{!}{%
\begin{tabular}{c|
S[table-format=2.2(1)]
S[table-format=2.2(1)]
S[table-format=2.2(1)]
S[table-format=2.2(1)]
}
\toprule
      & {STEM} &  {Humanities}  & {Social Sciences} & {Other} \\ \midrule
Word Influence   & 31.46\textsubscript{0.54} &  27.90\textsubscript{1.77} &  37.09\textsubscript{0.79} & 46.20\textsubscript{0.40} \\ 
{Random}  & 30.32\textsubscript{0.88}  &  25.48\textsubscript{1.64}  &  36.20\textsubscript{1.29}  &  44.69\textsubscript{1.14} \\ \bottomrule

\end{tabular}%
}
\caption{Ablation results on the search method. Random denotes searching on a random order of words.}
\label{main-ablation-random}
\vspace{-2mm}
\end{table}

\begin{figure}[!tbp]
    \centering
  \includegraphics[width=1.0\columnwidth]{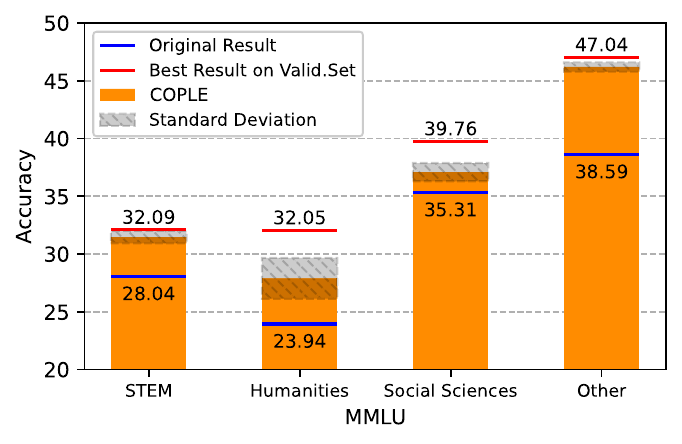}
  \caption{Performance difference when COPLE is performed on proxy reference tasks and on validation set.}
  \label{compare_with_optimal}
\end{figure}

\paragraph{How far is COPLE from the optimal results?}

Figure~\ref{compare_with_optimal} shows the performance gap between the potential best prompts found by COPLE directly on the validation set (\ref{combinatorial_optimization}) and on proxy tasks (\ref{reference_optimization}). The results show that the proxy tasks provide a reasonable approximation of the target task distribution, and the performance of COPLE is close to optimal.

\section{Conclusion}

In this paper, we demonstrate the notable lexical sensitivity of LLMs to prompts, which potentially degrades their performance on downstream tasks. We show that even semantically similar prompts located in the neighborhood of the latent representation space may yield very different results. To recover the performance drop caused by the sensitivity, we propose COPLE, a black-box combinatorial optimization framework that iteratively improves lexical choices in prompts. Experiments illustrate the effectiveness of COPLE in recovering both the model's ability of instruct-following and solving downstream tasks. We believe that carefully checking the word usage is essential before performing complex prompt engineering.

\clearpage

\section*{Limitations}

Despite the effectiveness of COPLE, we want to discuss some limitations of this work. Firstly, our experimental scope is primarily restricted to models around the 7-billion-parameter scale, as our computational resources are limited. Secondly, while we focus on optimizing the lexical choices within the task description component of the prompts, it is possible that lexical sensitivity affects the entirety of a prompt. However, expanding our optimization to include the full prompt significantly increases the size of search space, making the experiment computationally infeasible with our current resources. Thirdly, although we believe that lexical optimization should be a fundamental step prior to more complex prompt engineering methods, our research does not explore the combination of our proposed COPLE framework with other prompt engineering strategies that potentially yield further improvements in model performance. Despite these limitations, our study provides insights into the influence of lexical variation on language model prompts, from both the perspective of downstream performance and latent sentence representation. The findings highlight that even subtle lexical changes, when systematically optimized, can significantly enhance the performance of language models on downstream tasks.

\section*{Acknowledgements}
This work was supported by the Youth Innovation Promotion Association CAS (No.2023166), and the Informatization Plan of Chinese Academy of Sciences (Grant No. CAS-WX2021SF-0508).

\bibliography{custom_sum}

\clearpage

\onecolumn

\appendix

\section{Additional Experimental Details}
\label{appendix_additional_details}

\subsection{Details on Dataset}
\label{appendix_dataset}

The General Language Understanding Evaluation (GLUE)~\cite{Wang2019GLUE} benchmark is a collection of datasets for training, evaluating, and analyzing natural language understanding systems. The subset used in our experiment include: (1) The Stanford Sentiment Treebank (SST2)~\cite{Socher2013SST2} consists of movie review sentences annotated for sentiment. (2) The Corpus of Linguistic Acceptability (CoLA)~\cite{Warstadt2019COLA} contains English acceptability judgments drawn from linguistic theory publications. (3) The Multi-Genre Natural Language Inference (MNLI)~\cite{Williams2018MNLI} corpus includes sentence pairs annotated with textual entailment information. (4) The Question-answering NLI (QNLI)~\cite{Rajpurkar2016QNLI} is derived from SQuAD, converted to a binary sentence pair classification task. (5) The Recognizing Textual Entailment (RTE) datasets come from a series of textual entailment challenges~\cite{dagan2005RTE1,bar2006RTE2,Giampiccolo2007RTE,bentivogli2009RTE5}. (6) The Microsoft Research Paraphrase Corpus (MRPC)~\cite{Dolan2005MRPC} contains sentence pairs annotated for semantic equivalence. (7) The Quora Question Pairs2 (QQP)~\cite{Cer2017QQP} includes question pairs from Quora annotated for semantic equivalence. The Massive Multitask Language Understanding (MMLU) dataset~\cite{Hendrycks2021MMLU} contains multiple-choice questions that cover 57 tasks, which can be divided into four main subsets: STEM, Humanities, Social Sciences, and Other, with 14,042 test and 1,531 validation examples. In our experiment, for constructing the proxy reference tasks, we sample from the test set on MMLU. More information about the datasets is provided in Table~\ref{table_appendix_dataset}.

\begin{table}[htbp]
\centering
\resizebox{0.9\textwidth}{!}{%
\begin{tabular}{
  c
  S[table-format=6.0]
  S[table-format=5.0]
  c
  c
}
\toprule
Dataset & {$\#$.Training example} & {$\#$.Validation example} & {Mission Type} & {$\#$.Category} \\ \midrule
SST2 & 67349 & 872 & Sentiment Analysis & 2 \\
CoLA & 8551 & 1043 & Linguistic Acceptability & 2 \\
MNLI & 392702 & 19647 & Natural Language Inference & 3 \\
QNLI & 104743 & 5463 & Natural Language Inference & 2 \\
RTE & 2490 & 277 & Natural Language Inference & 2 \\
MRPC & 3668 & 408 & Semantic Equivalence & 2 \\
QQP & 363849 & 40430 & Semantic Equivalence & 2 \\
MMLU & {14042 (Test)} & 1531 & Question Answering & 4 \\ 
\bottomrule
\end{tabular}
}
\caption{Summary of datasets used in the experiments. For MNLI, 19,647 validation examples consists of 9,815 from the matched in-domain section and 9,832 from the mismatched cross-domain section. For MMLU, we report the size of test set rather than training set.}
\label{table_appendix_dataset}
\vspace{-2mm}
\end{table}

\subsection{Details on Evaluation}
\label{appendix_evaluation}

We follow the same evaluation style as HELM~\cite{Lee2023HELM}, which expects the model to output the correct label word (for GLUE) or the letter of the correct option (for MMLU), rather than directly using the probability of the output token for judgement. For example, on SST2, for a sentence with positive sentiment, we expect the model output to be "\emph{positive}". Similarly, on MMLU, the model is expected to output one of the letters "\emph{A}", "\emph{B}", "\emph{C}", or "\emph{D}" that matches the correct answer. Otherwise, we consider the model makes incorrect decisions. Note that this evaluation approach may result in model performance worse than that of random guessing. However, we believe that it provides a more accurate indication of the model's ability on instruction following and on solving downstream tasks. For all datasets, we report the performance with Accuracy.

\subsection{Details on Baseline Prompts}
\label{appendix_prompt}

In the main text, we perform COPLE on various scenarios, including \emph{Original}, \emph{1-shot}, \emph{3-shot}, \emph{EP02}, \emph{EP03}, and \emph{Zero-shot-CoT}; here we report the detailed prompts of these scenarios. The initial prompts for GLUE are in Table~\ref{appendix_prompt_baseline_sst2}-\ref{appendix_prompt_baseline_qqp}, and the initial prompts for MMLU are in Table~\ref{appendix_prompt_baseline_mmlu_1}-\ref{appendix_prompt_baseline_mmlu_2}. It should also be noted that for scenarios of \emph{Emotion Prompts}, following \citet{Li2023EmotionPrompt}, we insert the additional text at the end of the task description and verbalizer, but before the demo examples, and we do not consider the additional text to be a part of the task descriptions to perform optimization.

\renewcommand{\arraystretch}{1.2}
\begin{table}[htbp]
\small
\centering
\resizebox{\textwidth}{!}{%
\begin{tabularx}{\textwidth}{c c >{\raggedright\arraybackslash}m{0.7\textwidth}}
\toprule
\textbf{Dataset} & \textbf{Scenario} & \textbf{Prompt} \\
\midrule
\multirow{6}{*}[-19em]{SST2} & Original &  \brtext{For the given sentence, label the sentiment of the sentence as positive or negative.} \grtext{Do not respond with anything other than the labels `positive' or `negative'.} \newline \newline Question: \{content\} \newline Answer:  \\
\cmidrule(l){2-3}
 & 1-shot &  \brtext{For the given sentence, label the sentiment of the sentence as positive or negative.} \grtext{Do not respond with anything other than the labels `positive' or `negative'.} \newline \newline \bltext{Question: \{demo\_content\} \newline Answer: \{demo\_answer\}} \newline \newline Question: \{content\} \newline Answer:  \\
\cmidrule(l){2-3}
 & 3-shot &  \brtext{For the given sentence, label the sentiment of the sentence as positive or negative.} \grtext{Do not respond with anything other than the labels `positive' or `negative'.} \newline \newline \bltext{Question: \{demo\_content\_1\} \newline Answer: \{demo\_answer\_1\} \newline \newline Question: \{demo\_content\_2\} \newline Answer: \{demo\_answer\_2\} \newline \newline Question: \{demo\_content\_3\} \newline Answer: \{demo\_answer\_3\}} \newline \newline Question: \{content\} \newline Answer:  \\
\cmidrule(l){2-3}
 & EP02 &  \brtext{For the given sentence, label the sentiment of the sentence as positive or negative.} \grtext{Do not respond with anything other than the labels `positive' or `negative'.} This is very important to my career. \newline \newline Question: \{content\} \newline Answer:  \\
\cmidrule(l){2-3}
 & EP03 &  \brtext{For the given sentence, label the sentiment of the sentence as positive or negative.} \grtext{Do not respond with anything other than the labels `positive' or `negative'.} You'd better be sure. \newline \newline Question: \{content\} \newline Answer:  \\
\cmidrule(l){2-3}
 & Zero-shot CoT &  \brtext{For the given sentence, label the sentiment of the sentence as positive or negative.} Let's think step by step. \grtext{Then, end the response with \char`\"Therefore, the answer is: <label>`positive' / `negative'</label>.\char`\"} \newline \newline Question: \{content\} \newline Answer:  \\
\bottomrule
\end{tabularx}
}
\caption{Detailed baseline prompts for SST2. In the prompt, \brtext{brown text} indicates the task description, which is the target that COPLE performs on, \grtext{green text} indicates the verbalizer, and \bltext{blue text} indicates the demo examples for few-shot learning. ``\emph{\{text\}}'' denotes the placeholder, which will be replaced with the text sampled from the dataset.}
\label{appendix_prompt_baseline_sst2}
\end{table}

\renewcommand{\arraystretch}{1.2}
\begin{table}[htbp]
\small
\centering
\resizebox{\textwidth}{!}{%
\begin{tabularx}{\textwidth}{c c >{\raggedright\arraybackslash}m{0.7\textwidth}}
\toprule
\textbf{Dataset} & \textbf{Scenario} & \textbf{Prompt} \\
\midrule
\multirow{6}{*}[-19em]{CoLA} & Original & \brtext{Does this sentence make sense?} \grtext{Do not respond with anything other than the labels `Yes' or `No'.}  \newline \newline Question: \{content\} \newline Answer:  \\
\cmidrule(l){2-3}
 & 1-shot & \brtext{Does this sentence make sense?} \grtext{Do not respond with anything other than the labels `Yes' or `No'.} \newline \newline \bltext{Question: \{demo\_content\} \newline Answer: \{demo\_answer\}} \newline \newline Question: \{content\} \newline Answer:  \\
\cmidrule(l){2-3}
 & 3-shot & \brtext{Does this sentence make sense?} \grtext{Do not respond with anything other than the labels `Yes' or `No'.} \newline \newline  \bltext{Question: \{demo\_content\_1\} \newline Answer: \{demo\_answer\_1\} \newline \newline Question: \{demo\_content\_2\} \newline Answer: \{demo\_answer\_2\} \newline \newline Question: \{demo\_content\_3\} \newline Answer: \{demo\_answer\_3\}} \newline \newline Question: \{content\} \newline Answer:  \\
\cmidrule(l){2-3}
 & EP02 & \brtext{Does this sentence make sense?} \grtext{Do not respond with anything other than the labels `Yes' or `No'.} This is very important to my career. \newline \newline Question: \{content\} \newline Answer:  \\
\cmidrule(l){2-3}
 & EP03 & \brtext{Does this sentence make sense?} \grtext{Do not respond with anything other than the labels `Yes' or `No'.} You'd better be sure. \newline \newline Question: \{content\} \newline Answer: \\
\cmidrule(l){2-3}
 & Zero-shot CoT & \brtext{Does this sentence make sense?} Let's think step by step. \grtext{Then, end the response with \char`\"Therefore, the answer is: <label>`Yes' / `No'</label>.\char`\"} \newline \newline Question: \{content\} \newline Answer:  \\
\bottomrule
\end{tabularx}
}
\caption{Detailed baseline prompts for CoLA. In the prompt, \brtext{brown text} indicates the task description, which is the target that COPLE performs on, \grtext{green text} indicates the verbalizer, and \bltext{blue text} indicates the demo examples for few-shot learning. ``\emph{\{text\}}'' denotes the placeholder, which will be replaced with the text sampled from the dataset.}
\label{appendix_prompt_baseline_cola}
\end{table}

\renewcommand{\arraystretch}{1.2}
\begin{table}[htbp]
\small
\centering
\resizebox{\textwidth}{!}{%
\begin{tabularx}{\textwidth}{c c >{\raggedright\arraybackslash}m{0.7\textwidth}}
\toprule
\textbf{Dataset} & \textbf{Scenario} & \textbf{Prompt} \\
\midrule
\multirow{6}{*}[-19em]{MNLI} & Original & \brtext{Please identify whether the premise entails the hypothesis.} \grtext{Do not respond with anything other than the labels `entailment', `neutral', or `contradiction'.} \newline \newline Question: \{content\} \newline Answer:  \\
\cmidrule(l){2-3}
 & 1-shot & \brtext{Please identify whether the premise entails the hypothesis.} \grtext{Do not respond with anything other than the labels `entailment', `neutral', or `contradiction'.} \newline \newline \bltext{Question: \{demo\_content\} \newline Answer: \{demo\_answer\}} \newline \newline Question: \{content\} \newline Answer:  \\
\cmidrule(l){2-3}
 & 3-shot & \brtext{Please identify whether the premise entails the hypothesis.} \grtext{Do not respond with anything other than the labels `entailment', `neutral', or `contradiction'.} \newline \newline \bltext{Question: \{demo\_content\_1\} \newline Answer: \{demo\_answer\_1\} \newline \newline Question: \{demo\_content\_2\} \newline Answer: \{demo\_answer\_2\} \newline \newline Question: \{demo\_content\_3\} \newline Answer: \{demo\_answer\_3\}} \newline \newline Question: \{content\} \newline Answer:  \\
\cmidrule(l){2-3}
 & EP02 & \brtext{Please identify whether the premise entails the hypothesis.} \grtext{Do not respond with anything other than the labels `entailment', `neutral', or `contradiction'.} This is very important to my career. \newline \newline Question: \{content\} \newline Answer:  \\
\cmidrule(l){2-3}
 & EP03 & \brtext{Please identify whether the premise entails the hypothesis.} \grtext{Do not respond with anything other than the labels `entailment', `neutral', or `contradiction'.} You'd better be sure. \newline \newline Question: \{content\} \newline Answer:  \\
\cmidrule(l){2-3}
 & Zero-shot CoT & \brtext{Please identify whether the premise entails the hypothesis.} Let's think step by step. \grtext{Then, end the response with \char`\"Therefore, the answer is: <label>`entailment' / `neutral' / `contradiction'</label>. \char`\"} \newline \newline Question: \{content\} \newline Answer:  \\
\bottomrule
\end{tabularx}
}
\caption{Detailed baseline prompts for MNLI. In the prompt, \brtext{brown text} indicates the task description, which is the target that COPLE performs on, \grtext{green text} indicates the verbalizer, and \bltext{blue text} indicates the demo examples for few-shot learning. ``\emph{\{text\}}'' denotes the placeholder, which will be replaced with the text sampled from the dataset.}
\label{appendix_prompt_baseline_mnli}
\end{table}

\renewcommand{\arraystretch}{1.2}
\begin{table}[htbp]
\small
\centering
\resizebox{\textwidth}{!}{%
\begin{tabularx}{\textwidth}{c c >{\raggedright\arraybackslash}m{0.7\textwidth}}
\toprule
\textbf{Dataset} & \textbf{Scenario} & \textbf{Prompt} \\
\midrule
\multirow{6}{*}[-19em]{QNLI} & Original & \brtext{Please identify whether the sentence answers the question.} \grtext{Do not respond with anything other than the labels `Yes' or `No'.}  \newline \newline Question: \{content\} \newline Answer:  \\
\cmidrule(l){2-3}
 & 1-shot & \brtext{Please identify whether the sentence answers the question.} \grtext{Do not respond with anything other than the labels `Yes' or `No'.} \newline \newline \bltext{Question: \{demo\_content\} \newline Answer: \{demo\_answer\}} \newline \newline Question: \{content\} \newline Answer:  \\
\cmidrule(l){2-3}
 & 3-shot & \brtext{Please identify whether the sentence answers the question.} \grtext{Do not respond with anything other than the labels `Yes' or `No'.} \newline \newline  \bltext{Question: \{demo\_content\_1\} \newline Answer: \{demo\_answer\_1\} \newline \newline Question: \{demo\_content\_2\} \newline Answer: \{demo\_answer\_2\} \newline \newline Question: \{demo\_content\_3\} \newline Answer: \{demo\_answer\_3\}} \newline \newline Question: \{content\} \newline Answer:  \\
\cmidrule(l){2-3}
 & EP02 & \brtext{Please identify whether the sentence answers the question.} \grtext{Do not respond with anything other than the labels `Yes' or `No'.} This is very important to my career. \newline \newline Question: \{content\} \newline Answer:  \\
\cmidrule(l){2-3}
 & EP03 & \brtext{Please identify whether the sentence answers the question.} \grtext{Do not respond with anything other than the labels `Yes' or `No'.} You'd better be sure. \newline \newline Question: \{content\} \newline Answer:  \\
\cmidrule(l){2-3}
 & Zero-shot CoT & \brtext{Please identify whether the sentence answers the question.} Let's think step by step. \grtext{Then, end the response with \char`\"Therefore, the answer is: <label>`Yes' / `No'</label>.\char`\"} \newline \newline Question: \{content\} \newline Answer:  \\
\bottomrule
\end{tabularx}
}
\caption{Detailed baseline prompts for QNLI. In the prompt, \brtext{brown text} indicates the task description, which is the target that COPLE performs on, \grtext{green text} indicates the verbalizer, and \bltext{blue text} indicates the demo examples for few-shot learning. ``\emph{\{text\}}'' denotes the placeholder, which will be replaced with the text sampled from the dataset.}
\label{appendix_prompt_baseline_qnli}
\end{table}

\renewcommand{\arraystretch}{1.2}
\begin{table}[htbp]
\small
\centering
\resizebox{\textwidth}{!}{%
\begin{tabularx}{\textwidth}{c c >{\raggedright\arraybackslash}m{0.7\textwidth}}
\toprule
\textbf{Dataset} & \textbf{Scenario} & \textbf{Prompt} \\
\midrule
\multirow{6}{*}[-19em]{RTE} & Original & \brtext{Please identify whether the premise entails the hypothesis.} \grtext{Do not respond with anything other than the labels `Yes' or `No'.}  \newline \newline Question: \{content\} \newline Answer:  \\
\cmidrule(l){2-3}
 & 1-shot & \brtext{Please identify whether the premise entails the hypothesis.} \grtext{Do not respond with anything other than the labels `Yes' or `No'.} \newline \newline \bltext{Question: \{demo\_content\} \newline Answer: \{demo\_answer\}} \newline \newline Question: \{content\} \newline Answer:  \\
\cmidrule(l){2-3}
 & 3-shot & \brtext{Please identify whether the premise entails the hypothesis.} \grtext{Do not respond with anything other than the labels `Yes' or `No'.} \newline \newline  \bltext{Question: \{demo\_content\_1\} \newline Answer: \{demo\_answer\_1\} \newline \newline Question: \{demo\_content\_2\} \newline Answer: \{demo\_answer\_2\} \newline \newline Question: \{demo\_content\_3\} \newline Answer: \{demo\_answer\_3\}} \newline \newline Question: \{content\} \newline Answer:  \\
\cmidrule(l){2-3}
 & EP02 & \brtext{Please identify whether the premise entails the hypothesis.} \grtext{Do not respond with anything other than the labels `Yes' or `No'.} This is very important to my career. \newline\newline  Question: \{content\} \newline Answer:  \\
\cmidrule(l){2-3}
 & EP03 & \brtext{Please identify whether the premise entails the hypothesis.} \grtext{Do not respond with anything other than the labels `Yes' or `No'.} You'd better be sure. \newline \newline Question: \{content\} \newline Answer:  \\
\cmidrule(l){2-3}
 & Zero-shot CoT & \brtext{Please identify whether the premise entails the hypothesis.} Let's think step by step. \grtext{Then, end the response with \char`\"Therefore, the answer is: <label>`Yes' / `No'</label>.\char`\"} \newline \newline Question: \{content\} \newline Answer:  \\
\bottomrule
\end{tabularx}
}
\caption{Detailed baseline prompts for RTE. In the prompt, \brtext{brown text} indicates the task description, which is the target that COPLE performs on, \grtext{green text} indicates the verbalizer, and \bltext{blue text} indicates the demo examples for few-shot learning. ``\emph{\{text\}}'' denotes the placeholder, which will be replaced with the text sampled from the dataset.}
\label{appendix_prompt_baseline_rte}
\end{table}

\renewcommand{\arraystretch}{1.2}
\begin{table}[htbp]
\small
\centering
\resizebox{\textwidth}{!}{%
\begin{tabularx}{\textwidth}{c c >{\raggedright\arraybackslash}m{0.7\textwidth}}
\toprule
\textbf{Dataset} & \textbf{Scenario} & \textbf{Prompt} \\
\midrule
\multirow{6}{*}[-19em]{MRPC} & Original & \brtext{Do both sentences mean the same thing?} \grtext{Do not respond with anything other than the labels `Yes' or `No'.}  \newline \newline Question: \{content\} \newline Answer:  \\
\cmidrule(l){2-3}
 & 1-shot & \brtext{Do both sentences mean the same thing?} \grtext{Do not respond with anything other than the labels `Yes' or `No'.} \newline \newline \bltext{Question: \{demo\_content\} \newline Answer: \{demo\_answer\}} \newline \newline Question: \{content\} \newline Answer:  \\
\cmidrule(l){2-3}
 & 3-shot & \brtext{Do both sentences mean the same thing?} \grtext{Do not respond with anything other than the labels `Yes' or `No'.} \newline \newline  \bltext{Question: \{demo\_content\_1\} \newline Answer: \{demo\_answer\_1\} \newline \newline Question: \{demo\_content\_2\} \newline Answer: \{demo\_answer\_2\} \newline \newline Question: \{demo\_content\_3\} \newline Answer: \{demo\_answer\_3\}} \newline \newline Question: \{content\} \newline Answer:  \\
\cmidrule(l){2-3}
 & EP02 & \brtext{Do both sentences mean the same thing?} \grtext{Do not respond with anything other than the labels `Yes' or `No'.} This is very important to my career. \newline \newline Question: \{content\} \newline Answer:  \\
\cmidrule(l){2-3}
 & EP03 & \brtext{Do both sentences mean the same thing?} \grtext{Do not respond with anything other than the labels `Yes' or `No'.} You'd better be sure. \newline \newline Question: \{content\} \newline Answer:  \\
\cmidrule(l){2-3}
 & Zero-shot CoT & \brtext{Do both sentences mean the same thing?} Let's think step by step. \grtext{Then, end the response with \char`\"Therefore, the answer is: <label>`Yes' / `No'</label>.\char`\"} \newline \newline Question: \{content\} \newline Answer:  \\
\bottomrule
\end{tabularx}
}
\caption{Detailed baseline prompts for MRPC. In the prompt, \brtext{brown text} indicates the task description, which is the target that COPLE performs on, \grtext{green text} indicates the verbalizer, and \bltext{blue text} indicates the demo examples for few-shot learning. ``\emph{\{text\}}'' denotes the placeholder, which will be replaced with the text sampled from the dataset.}
\label{appendix_prompt_baseline_mrpc}
\end{table}

\renewcommand{\arraystretch}{1.2}
\begin{table}[htbp]
\small
\centering
\resizebox{\textwidth}{!}{%
\begin{tabularx}{\textwidth}{c c >{\raggedright\arraybackslash}m{0.7\textwidth}}
\toprule
\textbf{Dataset} & \textbf{Scenario} & \textbf{Prompt} \\
\midrule
\multirow{6}{*}[-19em]{QQP} & Original & \brtext{Please identify whether the sentences have the same meaning.} \grtext{Do not respond with anything other than the labels `equal' or `unequal'.}  \newline \newline Question: \{content\} \newline Answer:  \\
\cmidrule(l){2-3}
 & 1-shot & \brtext{Please identify whether the sentences have the same meaning.} \grtext{Do not respond with anything other than the labels `equal' or `unequal'.} \newline \newline \bltext{Question: \{demo\_content\} \newline Answer: \{demo\_answer\}} \newline \newline Question: \{content\} \newline Answer:  \\
\cmidrule(l){2-3}
 & 3-shot & \brtext{Please identify whether the sentences have the same meaning.} \grtext{Do not respond with anything other than the labels `equal' or `unequal'.} \newline \newline  \bltext{Question: \{demo\_content\_1\} \newline Answer: \{demo\_answer\_1\} \newline \newline Question: \{demo\_content\_2\} \newline Answer: \{demo\_answer\_2\} \newline \newline Question: \{demo\_content\_3\} \newline Answer: \{demo\_answer\_3\}} \newline \newline Question: \{content\} \newline Answer:  \\
\cmidrule(l){2-3}
 & EP02 & \brtext{Please identify whether the sentences have the same meaning.} \grtext{Do not respond with anything other than the labels `equal' or `unequal'.} This is very important to my career. \newline \newline Question: \{content\} \newline Answer:  \\
\cmidrule(l){2-3}
 & EP03 & \brtext{Please identify whether the sentences have the same meaning.} \grtext{Do not respond with anything other than the labels `equal' or `unequal'.} You'd better be sure. \newline \newline Question: \{content\} \newline Answer:  \\
\cmidrule(l){2-3}
 & Zero-shot CoT & \brtext{Please identify whether the sentences have the same meaning.} Let's think step by step. \grtext{Then, end the response with \char`\"Therefore, the answer is: <label>`equal' / `unequal'</label>.\char`\"} \newline \newline Question: \{content\} \newline Answer:  \\
\bottomrule
\end{tabularx}
}
\caption{Detailed baseline prompts for QQP. In the prompt, \brtext{brown text} indicates the task description, which is the target that COPLE performs on, \grtext{green text} indicates the verbalizer, and \bltext{blue text} indicates the demo examples for few-shot learning. ``\emph{\{text\}}'' denotes the placeholder, which will be replaced with the text sampled from the dataset.}
\label{appendix_prompt_baseline_qqp}
\end{table}

\renewcommand{\arraystretch}{1.0}
\begin{table}[htbp]
\small
\centering
\resizebox{0.95\textwidth}{!}{%
\begin{tabularx}{\textwidth}{c c >{\raggedright\arraybackslash}m{0.7\textwidth}}
\toprule
\textbf{Dataset} & \textbf{Scenario} & \textbf{Prompt} \\
\midrule
\multirow{6}{*}[-24em]{MMLU} & Original & \brtext{The following are multiple choice questions (with answers) about \{task\}.} \grtext{Do not respond with anything other than the answer labels `A', `B', `C', or `D'.} \newline \newline Question: \{question\} \newline A. \{option\_A\} \newline B. \{option\_B\} \newline C. \{option\_C\} \newline D. \{option\_D\} \newline \newline Answer:  \\
\cmidrule(l){2-3}
 & 1-shot & \brtext{The following are multiple choice questions (with answers) about \{task\}.} \grtext{Do not respond with anything other than the answer labels `A', `B', `C', or `D'.} \newline \newline  \bltext{Question: \{demo\_question\} \newline A. \{demo\_option\_A\} \newline B. \{demo\_option\_B\} \newline C. \{demo\_option\_C\} \newline D. \{demo\_option\_D\} \newline \newline Answer: \{demo\_answer\}} \newline \newline Question: \{question\} \newline A. \{option\_A\} \newline B. \{option\_B\} \newline C. \{option\_C\} \newline D. \{option\_D\} \newline \newline Answer:  \\
\cmidrule(l){2-3}
 & 3-shot & \brtext{The following are multiple choice questions (with answers) about \{task\}.} \grtext{Do not respond with anything other than the answer labels `A', `B', `C', or `D'.} \newline \newline  \bltext{Question: \{demo\_question\_1\} \newline A. \{demo\_option\_A\_1\} \newline B. \{demo\_option\_B\_1\} \newline C. \{demo\_option\_C\_1\} \newline D. \{demo\_option\_D\_1\} \newline \newline Answer: \{demo\_answer\_1\} \newline \newline Question: \{demo\_question\_2\} \newline A. \{demo\_option\_A\_2\} \newline B. \{demo\_option\_B\_2\} \newline C. \{demo\_option\_C\_2\} \newline D. \{demo\_option\_D\_2\} \newline \newline Answer: \{demo\_answer\_2\} \newline \newline Question: \{demo\_question\_3\} \newline A. \{demo\_option\_A\_3\} \newline B. \{demo\_option\_B\_3\} \newline C. \{demo\_option\_C\_3\} \newline D. \{demo\_option\_D\_3\} \newline \newline Answer: \{demo\_answer\_3\}} \newline \newline Question: \{question\} \newline A. \{option\_A\} \newline B. \{option\_B\} \newline C. \{option\_C\} \newline D. \{option\_D\} \newline \newline Answer:  \\
\bottomrule
\end{tabularx}
}
\caption{Detailed baseline prompts of \emph{Original} and \emph{In-context Learning} setting for MMLU. In the prompt, \brtext{brown text} indicates the task description, which is the target that COPLE performs on, \grtext{green text} indicates the verbalizer, and \bltext{blue text} indicates the demo examples for few-shot learning. ``\emph{\{text\}}'' denotes the placeholder, which will be replaced with the text sampled from the dataset.}
\label{appendix_prompt_baseline_mmlu_1}
\end{table}

\renewcommand{\arraystretch}{1.0}
\begin{table}[!ht]
\small
\centering
\resizebox{1.0\textwidth}{!}{%
\begin{tabularx}{\textwidth}{c c >{\raggedright\arraybackslash}m{0.7\textwidth}}
\toprule
\textbf{Dataset} & \textbf{Scenario} & \textbf{Prompt} \\
\midrule
\multirow{6}{*}[-10em]{MMLU} & EP02 & \brtext{The following are multiple choice questions (with answers) about \{task\}.} \grtext{Do not respond with anything other than the answer labels `A', `B', `C', or `D'.} This is very important to my career. \newline \newline Question: \{question\} \newline A. \{option\_A\} \newline B. \{option\_B\} \newline C. \{option\_C\} \newline D. \{option\_D\} \newline \newline Answer:  \\
\cmidrule(l){2-3}
 & EP03 & \brtext{The following are multiple choice questions (with answers) about \{task\}.} \grtext{Do not respond with anything other than the answer labels `A', `B', `C', or `D'.} You'd better be sure. \newline \newline Question: \{question\} \newline A. \{option\_A\} \newline B. \{option\_B\} \newline C. \{option\_C\} \newline D. \{option\_D\} \newline \newline Answer: \\
\cmidrule(l){2-3}
 & Zero-shot CoT & \brtext{The following are multiple choice questions (with answers) about \{task\}.} Let's think step by step. \grtext{Then, end the response with \char`\"Therefore, the answer is: <label>`A' / `B' / `C' / `D'</label>.\char`\"} \newline \newline Question: \{question\} \newline A. \{option\_A\} \newline B. \{option\_B\} \newline C. \{option\_C\} \newline D. \{option\_D\} \newline \newline Answer:  \\
\bottomrule
\end{tabularx}
}
\caption{Detailed baseline prompts of \emph{Emotion Prompt} and \emph{Chain-of-thought} setting for MMLU. In the prompt, \brtext{brown text} indicates the task description, which is the target that COPLE performs on, \grtext{green text} indicates the verbalizer, and \bltext{blue text} indicates the demo examples for few-shot learning. ``\emph{\{text\}}'' denotes the placeholder, which will be replaced with the text sampled from the dataset.}
\label{appendix_prompt_baseline_mmlu_2}
\vspace{4mm}
\end{table}

\section{Additional Experimental Results}

\subsection{Details on the Optimized Prompt Crafted by COPLE}
\label{appendix_optimized_prompt}
Table~\ref{appendix_prompt_optimized_llama2_glue}-\ref{appendix_prompt_optimized_llama2_mmlu} shows the optimal task descriptions optimized by COPLE on \texttt{Llama2-7B-chat} for various scenarios. Table~\ref{appendix_prompt_optimized_mistral_glue}-\ref{appendix_prompt_optimized_mistral_mmlu} shows the optimal task descriptions optimized by COPLE on \texttt{Mistral-7B-Instruct-v0.1} for various scenarios. Table~\ref{appendix_prompt_optimized_gpt35_all} shows the optimal task description optimized by COPLE on ChatGPT (\texttt{gpt-3.5-turbo-0125}) and the \emph{Original} scenario.

\renewcommand{\arraystretch}{0.8}
\begin{table}[htbp]
\small
\centering
\resizebox{0.95\textwidth}{!}{%
\begin{tabularx}{\textwidth}{c c >{\hsize=0.5\hsize}X >{\hsize=1.5\hsize}X}
\toprule
\multirow{2}{*}{\textbf{Dataset}} & \multirow{2}{*}{\textbf{Scenario}} & \textbf{Task Description \newline (Before)} & \textbf{Task Description \newline (After)} \\
\midrule
\multirow{6}{*}[-5em]{SST2} & Original &  \multirow{6}{=}[-5em]{For the given sentence, label the sentiment of the sentence as positive or negative.} & Before the given sentence, label the sentiment of the user as positively or negative.  \\
\cmidrule(l){4-4}
 & 1-shot &  & For the closing sentence, label this rest of next sentence as positive or positive. \\
 \cmidrule(l){4-4}
 & 3-shot &  & For another given sentence, label the comment preceding no sentence as angry or negative. \\
 \cmidrule(l){4-4}
 & EP02 &  & Regarding ANY OTHER NEWS, label the sentiment of the sentence positive or negative. \\
 \cmidrule(l){4-4}
 & EP03 &  & If the given sentence, label the quality of the statement as positively or negative. \\
 \cmidrule(l){4-4}
 & Zero-shot CoT &  & For the introductory discussion, label the sentiment of the reviewer for positive or opposite. \\
 \cmidrule(l){1-4}
\multirow{6}{*}[-2em]{CoLA} & Original &  \multirow{6}{=}[-2em]{Does this sentence make sense?} & Did THIS sentence make?  \\
\cmidrule(l){4-4}
 & 1-shot &  & Does every word make Sense? \\
 \cmidrule(l){4-4}
 & 3-shot &  & Does my sentence any sense? \\
 \cmidrule(l){4-4}
 & EP02 &  & Has this now make sense? \\
 \cmidrule(l){4-4}
 & EP03 &  & Made THIS sentence make it? \\
 \cmidrule(l){4-4}
 & Zero-shot CoT &  & Should THAT sentence made sense? \\
  \cmidrule(l){1-4}
\multirow{6}{*}[-2em]{MNLI} & Original &  \multirow{6}{=}[-2em]{Please identify whether the premise entails the hypothesis.} & Please indicate whether the content entails valid conclusion.  \\
\cmidrule(l){4-4}
 & 1-shot &  & Please identify all the elements entails the hypothesis. \\
 \cmidrule(l){4-4}
 & 3-shot &  & Please identify whether the premise is the facts. \\
 \cmidrule(l){4-4}
 & EP02 &  & must identify whether the premise opposes the hypothesis. \\
 \cmidrule(l){4-4}
 & EP03 &  & Results evaluate Whether your data entails the hypothesis. \\
 \cmidrule(l){4-4}
 & Zero-shot CoT &  & Note whether the answer satisfies the evidence. \\
 \cmidrule(l){1-4}
\multirow{6}{*}[-2em]{QNLI} & Original &  \multirow{6}{=}[-2em]{Please identify whether the sentence answers the question.} & Please specify whether the above answers the question.  \\
\cmidrule(l){4-4}
 & 1-shot &  & Please comment whether sentence answers the question. \\
 \cmidrule(l){4-4}
 & 3-shot &  & We consider whether the sentence answers the question. \\
 \cmidrule(l){4-4}
 & EP02 &  & Please assess whether the above answered the error. \\
 \cmidrule(l){4-4}
 & EP03 &  & Please correct whether this answered the question. \\
 \cmidrule(l){4-4}
 & Zero-shot CoT &  & Please assess whether the sentence answers the queries. \\
  \cmidrule(l){1-4}
\multirow{6}{*}[-2em]{RTE} & Original &  \multirow{6}{=}[-2em]{Please identify whether the premise entails the hypothesis.} & First estimate whether premise entails each hypothesis. \\
\cmidrule(l){4-4}
 & 1-shot &  & Please confirm whether current knowledge advances the hypothesis. \\
 \cmidrule(l){4-4}
 & 3-shot &  & Now identify whether the correction entails the hypothesis.\\
 \cmidrule(l){4-4}
 & EP02 &  & First identify whether the premise confirms each hypothesis. \\
 \cmidrule(l){4-4}
 & EP03 &  & Please state whether the conclusion justifies the hypothesis. \\
 \cmidrule(l){4-4}
 & Zero-shot CoT &  & Please identify whether certain premise entails correct hypothesis. \\
  \cmidrule(l){1-4}
\multirow{6}{*}[-2em]{MRPC} & Original &  \multirow{6}{=}[-2em]{Do both sentences mean the same thing?} & Do both sentences mean a same thing? \\
\cmidrule(l){4-4}
 & 1-shot &  & Do both languages mean the same thing? \\
 \cmidrule(l){4-4}
 & 3-shot &  & Do both letters mean the same thing? \\
 \cmidrule(l){4-4}
 & EP02 &  & Do parallel sentences repeat the same thing? \\
 \cmidrule(l){4-4}
 & EP03 &  & Are both sentences mean the same thing? \\
 \cmidrule(l){4-4}
 & Zero-shot CoT &  & did both sentences convey the same thing? \\
   \cmidrule(l){1-4}
\multirow{6}{*}[-2em]{QQP} & Original &  \multirow{6}{=}[-2em]{Please identify whether the sentences have the same meaning.} & Please identify since the posts have the same text. \\
\cmidrule(l){4-4}
 & 1-shot &  & will identify Where unrelated sentences have nearly same significance. \\
 \cmidrule(l){4-4}
 & 3-shot &  & Please identify whether the sentences have the same ending. \\
 \cmidrule(l){4-4}
 & EP02 &  & Please determine are the have the same meaning. \\
 \cmidrule(l){4-4}
 & EP03 &  & Please identify both your works have the same subject. \\
 \cmidrule(l){4-4}
 & Zero-shot CoT &  & Please note whether adjacent sentences display the identical meaning. \\
\bottomrule
\end{tabularx}
}
\caption{Detailed optimized task descriptions on \texttt{Llama2-7B-chat} and GLUE.}
\label{appendix_prompt_optimized_llama2_glue}
\end{table}

\renewcommand{\arraystretch}{1.0}
\begin{table}[htbp]
\small
\centering
\resizebox{\textwidth}{!}{%
\begin{tabularx}{\textwidth}{c c >{\hsize=0.7\hsize}X >{\hsize=1.3\hsize}X}
\toprule
\multirow{2}{*}{\textbf{Dataset}} & \multirow{2}{*}{\textbf{Scenario}} & \textbf{Task Description \newline (Before)} & \textbf{Task Description \newline (After)} \\
\midrule
\multirow{6}{*}[-5em]{MMLU-STEM} & Original &  \multirow{24}{=}[-20em]{The following are multiple choice questions (with answers) about \{task\}.} & THE exercises contain easy choice questions (with answers) about \{task\}.  \\
\cmidrule(l){4-4}
 & 1-shot &  & The following generates multiple discussion questions (using arrows) about \{task\}. \\
 \cmidrule(l){4-4}
 & 3-shot &  & Se following are multiple popular questions (with answers) about \{task\}. \\
 \cmidrule(l){4-4}
 & EP02 &  & The examples list multiple choice questions (with labels) without \{task\}.  \\
 \cmidrule(l){4-4}
 & EP03 &  & The are infinite choice questions (by data) about \{task\}.  \\
 \cmidrule(l){4-4}
 & Zero-shot CoT &  & The following are multiple choice problems (default answers) involving \{task\}. \\
 \cmidrule(l){1-2} \cmidrule(l){4-4}
\multirow{6}{*}[-5em]{MMLU-Humanities} & Original &  & The model identifies multiple popular questions (which answers) about \{task\}  \\
\cmidrule(l){4-4}
 & 1-shot &  & In following are all survey questions (with answers) about \{task\}. \\
 \cmidrule(l){4-4}
 & 3-shot &  & The following are multiple questions (with answers) named \{task\}. \\
 \cmidrule(l){4-4}
 & EP02 &  & Graph following illustrates third choice answer (simple answers) about \{task\}. \\
 \cmidrule(l){4-4}
 & EP03 &  & The candidates are multiple choice questions (with) about \{task\}. \\
 \cmidrule(l){4-4}
 & Zero-shot CoT &  & The follows are most popular questions (mostly answers) about \{task\}. \\
  \cmidrule(l){1-2} \cmidrule(l){4-4}
\multirow{6}{*}[-5em]{MMLU-Social Sciences} & Original &  & The following displays multiple choice questions (with answers) about \{task\}. \\
\cmidrule(l){4-4}
 & 1-shot &  & The following multiple popular questions (with answers) about \{task\}. \\
 \cmidrule(l){4-4}
 & 3-shot &  & The following are multiple different questions (with answers) about \{task\}. \\
 \cmidrule(l){4-4}
 & EP02 &  & The following summarizes the choice questions (complete ones) about \{task\}.  \\
 \cmidrule(l){4-4}
 & EP03 &  & The entries present multiple choice questions (with answers) involving \{task\}. \\
 \cmidrule(l){4-4}
 & Zero-shot CoT &  & following are some choice questions (complete answers) about \{task\}. \\
 \cmidrule(l){1-2} \cmidrule(l){4-4}
\multirow{6}{*}[-5em]{MMLU-Other} & Original &  & The candidates posted multiple choice questions (with answers) about \{task\}.  \\
\cmidrule(l){4-4}
 & 1-shot &  & The following are choice questions (with answers) about \{task\}. \\
 \cmidrule(l){4-4}
 & 3-shot &  & The following are multiple standard answers (with answers) about \{task\}. \\
 \cmidrule(l){4-4}
 & EP02 &  & The Following are the choice questions (and answers) following \{task\}.  \\
 \cmidrule(l){4-4}
 & EP03 &  & The following implements multiple choice questions (No exceptions) about \{task\}.
\\
 \cmidrule(l){4-4}
 & Zero-shot CoT &  & The following contains your first questions (with explanation) about \{task\}. \\

\bottomrule
\end{tabularx}
}
\caption{Detailed optimized task descriptions on \texttt{Llama2-7B-chat} and MMLU. ``\emph{\{task\}}'' denotes the placeholder, which will be replaced with the detailed subset type.}
\label{appendix_prompt_optimized_llama2_mmlu}
\end{table}

\renewcommand{\arraystretch}{0.8}
\begin{table}[htbp]
\small
\centering
\resizebox{\textwidth}{!}{%
\begin{tabularx}{\textwidth}{c c >{\hsize=0.5\hsize}X >{\hsize=1.5\hsize}X}
\toprule
\multirow{2}{*}{\textbf{Dataset}} & \multirow{2}{*}{\textbf{Scenario}} & \textbf{Task Description \newline (Before)} & \textbf{Task Description \newline (After)} \\
\midrule
\multirow{6}{*}[-5em]{SST2} & Original &  \multirow{6}{=}[-5em]{For the given sentence, label the sentiment of the sentence as positive or negative.} & For the given sentence, label the sentiment of the sentence as positive or constructive.  \\
\cmidrule(l){4-4}
 & 1-shot &  & For the given sentence, keep positive sentiment of the sentence as positive not negative. \\
 \cmidrule(l){4-4}
 & 3-shot &  & Before the given sentence, express components of the sentence as positive OR negatively. \\
 \cmidrule(l){4-4}
 & EP02 &  & After the desired context, label every result of the process as happy or negative. \\
 \cmidrule(l){4-4}
 & EP03 &  & For my given question, label the sentiment of the sentence as positive or optimistic. \\
 \cmidrule(l){4-4}
 & Zero-shot CoT &  & For the given expression, label the outcome of the sentence as positive or. \\
 \cmidrule(l){1-4}
\multirow{6}{*}[-1.5em]{CoLA} & Original &  \multirow{6}{=}[-2em]{Does this sentence make sense?} & Does this sentence form follows?  \\
\cmidrule(l){4-4}
 & 1-shot &  & Does each word contain sense? \\
 \cmidrule(l){4-4}
 & 3-shot &  & Does each sentence make points? \\
 \cmidrule(l){4-4}
 & EP02 &  & Do I sentence make sense? \\
 \cmidrule(l){4-4}
 & EP03 &  & Does each sentence make sense? \\
 \cmidrule(l){4-4}
 & Zero-shot CoT &  & Has this sentence make sense? \\
  \cmidrule(l){1-4}
\multirow{6}{*}[-1.5em]{MNLI} & Original &  \multirow{6}{=}[-2em]{Please identify whether the premise entails the hypothesis.} & Please assess how the result entails proposed hypothesis.  \\
\cmidrule(l){4-4}
 & 1-shot &  & Please assess whether sufficient premise entails the claim. \\
 \cmidrule(l){4-4}
 & 3-shot &  & Please identify between and premise the hypothesis. \\
 \cmidrule(l){4-4}
 & EP02 &  & Please show whether the answer entails the hypothesis. \\
 \cmidrule(l){4-4}
 & EP03 &  & Please identify whether the result entails the hypothesis. \\
 \cmidrule(l){4-4}
 & Zero-shot CoT &  & Please evaluate whether the hypothesis entails my observations.  \\
 \cmidrule(l){1-4}
\multirow{6}{*}[-1.5em]{QNLI} & Original &  \multirow{6}{=}[-2em]{Please identify whether the sentence answers the question.} & Please identify whether any sentence asked either question.  \\
\cmidrule(l){4-4}
 & 1-shot &  & Please repeat in the affirmative regarding the question. \\
 \cmidrule(l){4-4}
 & 3-shot &  & please identify unless our article answers further question. \\
 \cmidrule(l){4-4}
 & EP02 &  & must identify whether the article answers the question. \\
 \cmidrule(l){4-4}
 & EP03 &  & Please identify whether the sentence addressed previous question. \\
 \cmidrule(l){4-4}
 & Zero-shot CoT &  & Please identify whether her sentence supports the question. \\
  \cmidrule(l){1-4}
\multirow{6}{*}[-1.5em]{RTE} & Original &  \multirow{6}{=}[-2em]{Please identify whether the premise entails the hypothesis.} & Please find sure the premise matches any hypothesis. \\
\cmidrule(l){4-4}
 & 1-shot &  & Please assess whether the premise supports the premises. \\
 \cmidrule(l){4-4}
 & 3-shot &  & Please identify HOW either premise fits any other.\\
 \cmidrule(l){4-4}
 & EP02 &  & Please verify either possible facts entails the claims. \\
 \cmidrule(l){4-4}
 & EP03 &  & Please verify when the claim matches the fact. \\
 \cmidrule(l){4-4}
 & Zero-shot CoT &  & Please identify that The premise entails the hypothesis. \\
  \cmidrule(l){1-4}
\multirow{6}{*}[-1.5em]{MRPC} & Original &  \multirow{6}{=}[-2em]{Do both sentences mean the same thing?} & Do both sentences suggest the same picture? \\
\cmidrule(l){4-4}
 & 1-shot &  & Do other sentences suggest a same theme? \\
 \cmidrule(l){4-4}
 & 3-shot &  & Do other sources suggest the Same thing? \\
 \cmidrule(l){4-4}
 & EP02 &  & both sentences mean the whole thing? \\
 \cmidrule(l){4-4}
 & EP03 &  & Do both sentences explain the same thing? \\
 \cmidrule(l){4-4}
 & Zero-shot CoT &  & Do both sentences describe the whole thing? \\
   \cmidrule(l){1-4}
\multirow{6}{*}[-2em]{QQP} & Original &  \multirow{6}{=}[-2em]{Please identify whether the sentences have the same meaning.} & Please identify whether following articles have the same keywords. \\
\cmidrule(l){4-4}
 & 1-shot &  & Please compare whether these sentences match the exact meaning. \\
 \cmidrule(l){4-4}
 & \multirow{2}{*}{3-shot} &  & Please identify whenever individual sentences have the equivalent content. \\
 \cmidrule(l){4-4}
 & EP02 &  & Please tell whether two quotes have the identical message. \\
 \cmidrule(l){4-4}
 & EP03 &  & Please identify whether the arguments have the same context. \\
 \cmidrule(l){4-4}
 & Zero-shot CoT &  & Please identify between the sentences have the same value. \\
\bottomrule
\end{tabularx}
}
\caption{Detailed optimized task descriptions on \texttt{Mistral-7B-Instruct-v0.1} and GLUE.}
\label{appendix_prompt_optimized_mistral_glue}
\end{table}

\renewcommand{\arraystretch}{1.0}
\begin{table}[htbp]
\small
\centering
\resizebox{\textwidth}{!}{%
\begin{tabularx}{\textwidth}{c  c >{\hsize=0.7\hsize}X >{\hsize=1.3\hsize}X}
\toprule
\multirow{2}{*}{\textbf{Dataset}} & \multirow{2}{*}{\textbf{Scenario}} & \textbf{Task Description \newline (Before)} & \textbf{Task Description \newline (After)} \\
\midrule
\multirow{14}{*}{MMLU-STEM} & \multirow{2}{*}{Original} &  \multirow{24}{=}[-20em]{The following are multiple choice questions (with answers) about \{task\}.} & The following are multiple choice questions (their variants) about \{task\}.  \\
\cmidrule(l){4-4}
 & \multirow{2}{*}{1-shot} &  & the following two first choice questions (with answers) about \{task\}. \\
 \cmidrule(l){4-4}
 & \multirow{2}{*}{3-shot} &  & Examples Here were multiple detailed questions (full comments) about \{task\}. \\
 \cmidrule(l){4-4}
 & \multirow{2}{*}{EP02} &  & The mes are multiple choice questions (with parentheses) about \{task\}.  \\
 \cmidrule(l){4-4}
 & \multirow{2}{*}{EP03} &  & Examples following are multiple example questions (complete answers) about \{task\}.  \\
 \cmidrule(l){4-4}
 & \multirow{2}{*}{Zero-shot CoT} &  & Then following are binary logic questions (with tags) about \{task\}. \\
 \cmidrule(l){1-2} \cmidrule(l){4-4}
\multirow{14}{*}{MMLU-Humanities} & \multirow{2}{*}{Original} &  & The are three choice cases (with answers) about \{task\}.   \\
\cmidrule(l){4-4}
 & \multirow{2}{*}{1-shot} &  & The Following are multiple choice questions (with hints) about \{task\}. \\
 \cmidrule(l){4-4}
 & \multirow{2}{*}{3-shot} &  & Then following are multiple choice questions (with answers) about \{task\}. \\
 \cmidrule(l){4-4}
 & \multirow{2}{*}{EP02} &  & The following are multiple obvious choices (easy fixes) using \{task\}. \\
 \cmidrule(l){4-4}
 & \multirow{2}{*}{EP03} &  & The Following are multiple choice messages (with answers) about \{task\}. \\
 \cmidrule(l){4-4}
 & \multirow{2}{*}{Zero-shot CoT} &  & The Below are multiple hypothetical questions (with answers) about \{task\}. \\
  \cmidrule(l){1-2} \cmidrule(l){4-4}
\multirow{14}{*}{MMLU-Social Sciences} & \multirow{2}{*}{Original} &  & The following are choice questions (with answers) about \{task\}. \\
\cmidrule(l){4-4}
 & \multirow{2}{*}{1-shot} &  & The following are ten sample questions (with keywords) about \{task\}. \\
 \cmidrule(l){4-4}
 & \multirow{2}{*}{3-shot} &  & The following are two sample questions (with answers) about \{task\}. \\
 \cmidrule(l){4-4}
 & \multirow{2}{*}{EP02} &  & The Following are multiple choice questions (with solutions) containing \{task\}.  \\
 \cmidrule(l){4-4}
 & \multirow{2}{*}{EP03} &  & The Following are one choice questions (with answers) about \{task\} \\
 \cmidrule(l){4-4}
 & \multirow{2}{*}{Zero-shot CoT} &  & The followed five multiple choice Questions (with Answers) about \{task\}. \\
 \cmidrule(l){1-2} \cmidrule(l){4-4}
\multirow{14}{*}{MMLU-Other} & \multirow{2}{*}{Original} &  & The following are multiple choice questions (answers) and \{task\}.  \\
\cmidrule(l){4-4}
 & \multirow{2}{*}{1-shot} &  & Then follow are two choice questions (with answers) about \{task\}. \\
 \cmidrule(l){4-4}
 & \multirow{2}{*}{3-shot} &  & Both following are multiple boolean questions (with answers) about \{task\}. \\
 \cmidrule(l){4-4}
 & \multirow{2}{*}{EP02} &  & The answers are Multiple choice questions (with answers) and \{task\}.  \\
 \cmidrule(l){4-4}
 & \multirow{2}{*}{EP03} &  & The following are multiple nested questions (with answers) about \{task\}.
\\
 \cmidrule(l){4-4}
 & \multirow{2}{*}{Zero-shot CoT} &  & Ch following is multiple choice questions (with answers) about \{task\}. \\

\bottomrule
\end{tabularx}
}
\caption{Detailed optimized task descriptions on \texttt{Mistral-7B-Instruct-v0.1} and MMLU. ``\emph{\{task\}}'' denotes the placeholder, which will be replaced with the detailed subset type.}
\label{appendix_prompt_optimized_mistral_mmlu}
\end{table}

\renewcommand{\arraystretch}{1.2}
\begin{table}[htbp]
\small
\centering
\resizebox{\textwidth}{!}{%
\begin{tabularx}{\textwidth}{c | >{\hsize=1\hsize}X | >{\hsize=1\hsize}X}
\toprule
\multirow{2}{*}{\textbf{Dataset}} & \textbf{Task Description \newline (Before)} & \textbf{Task Description \newline (After)} \\
\midrule
\multirow{2}{*}{SST2} & {For the given sentence, label the sentiment of the sentence as positive or negative.} & For the given article, label the sentiment above the sentence as positive or negative.  \\ \cmidrule(l){2-3}  
\multirow{1}{*}{CoLA} & {Does this sentence make sense?} & this sentence make sense?   \\ \cmidrule(l){2-3}  
\multirow{2}{*}{RTE} & {Please identify whether the premise entails the hypothesis.} & Please evidence the premise support current hypothesis.  \\ \cmidrule(l){2-3} 
\multirow{1}{*}{MRPC} & {Do both sentences mean the same thing?} & Do both answers mean this correct thing?  \\ \cmidrule(l){2-3}  
\multirow{2}{*}{MMLU-STEM} & \multirow{4}{=}[-2.5em]{The following are multiple choice questions (with answers) about \{task\}.} & The following are infinite choice questions (NO answers) within \{task\}.  \\   \cmidrule(l){3-3} 
\multirow{2}{*}{MMLU-Humanities} &  & Items above are a choice (with answers) about \{task\}.  \\   \cmidrule(l){3-3} 
\multirow{2}{*}{MMLU-Social Sciences} &  & The following answers multiple choice question (with answers) about \{task\}.  \\ \cmidrule(l){3-3}  
\multirow{2}{*}{MMLU-Other} & & The Following answers multiple choice questions (with answers) about \{task\}.  \\ \bottomrule
\end{tabularx}
}
\caption{Detailed optimized task descriptions on ChatGPT (\texttt{gpt-3.5-turbo-0125}) on the \emph{Original} scenario. ``\emph{\{task\}}'' denotes the placeholder, which will be replaced with the detailed subset type.}
\label{appendix_prompt_optimized_gpt35_all}
\end{table}

\end{document}